\documentclass{article}

\usepackage{microtype}
\usepackage{graphicx}
\usepackage{subfigure}
\usepackage{booktabs} 

\usepackage{hyperref}

\usepackage[accepted]{icml2024}

\usepackage{amsmath}
\usepackage{amssymb}
\usepackage{mathtools}
\usepackage{amsthm}

\usepackage[capitalize,noabbrev]{cleveref}

\theoremstyle{plain}
\newtheorem{theorem}{Theorem}[section]

\theoremstyle{definition}
\newtheorem{definition}[theorem]{Definition}

\theoremstyle{remark}

\usepackage[textsize=tiny]{todonotes}

\icmltitlerunning{Robust Stable Spiking Neural Networks}

\begin{document}

\twocolumn[
\icmltitle{Robust Stable Spiking Neural Networks}




\begin{icmlauthorlist}
\icmlauthor{Jianhao Ding}{pkucs}
\icmlauthor{Zhiyu Pan}{pkucs}
\icmlauthor{Yujia Liu}{pkucs}
\icmlauthor{Zhaofei Yu$^*$}{pkui,pkucs}
\icmlauthor{Tiejun Huang}{pkucs}
\end{icmlauthorlist}

\icmlaffiliation{pkucs}{School of Computer Science, Peking University, Beijing, China 100871}
\icmlaffiliation{pkui}{Institute for Artificial Intelligence, Peking University, Beijing, China 100871}

\icmlcorrespondingauthor{Zhaofei Yu}{yuzf12@pku.edu.cn}

\icmlkeywords{Machine Learning, ICML}

\vskip 0.3in
]



\printAffiliationsAndNotice{}  

\begin{abstract}
Spiking neural networks (SNNs) are gaining popularity in deep learning due to their low energy budget on neuromorphic hardware. However, they still face challenges in lacking sufficient robustness to guard safety-critical applications such as autonomous driving. Many studies have been conducted to defend SNNs from the threat of adversarial attacks. This paper aims to uncover the robustness of SNN through the lens of the stability of nonlinear systems. We are inspired by the fact that searching for parameters altering the leaky integrate-and-fire dynamics can enhance their robustness. Thus, we dive into the dynamics of membrane potential perturbation and simplify the formulation of the dynamics. We present that membrane potential perturbation dynamics can reliably convey the intensity of perturbation. Our theoretical analyses imply that the simplified perturbation dynamics satisfy input-output stability. Thus, we propose a training framework with modified SNN neurons and to reduce the mean square of membrane potential perturbation aiming at enhancing the robustness of SNN. Finally, we experimentally verify the effectiveness of the framework in the setting of Gaussian noise training and adversarial training on the image classification task. Please refer to \url{https://github.com/DingJianhao/stable-snn} for our code implementation.
\end{abstract}

\section{Introduction}
\label{introduction}


Spiking neural networks (SNNs) are gaining popularity in the field of deep learning, owing to their ability to deploy deep network architectures on neuromorphic hardware with high efficiency~\cite{pei2019towards,debole2019truenorth, davies2018loihi,nicolas2021sparse,fang2020spike}. Unlike typical Analog Neural Networks (ANNs), neurons in SNNs evolve their membrane potentials like biological systems in response to stimuli and use spike sequences to convey binary information~\cite{gerstner2014neuronal,xu2023constructing,zhu2024online}. This distinguishing feature distinguishes SNNs from ANNs, providing a simplified depiction of the complex dynamics observed in the brain~\cite{yao2022glif,zhang2020temporal,kheradpisheh2020temporal,shi2024spikingresformer}. Thus, training deep SNNs with good performance typically requires expanding the temporal evolution of SNNs due to their dynamic nature and employing surrogate functions to overcome the difficulty of the binary spike emission function~\cite{wu2018spatio,kim2020spiking,zhang2022rectified,rathi2021diet,kim2023exploring,xu2024enhancing,guo2023rmp}. The ongoing exploration of SNNs aims to bridge the computational capabilities of SNNs with the capabilities observed in biological neural systems, making them a promising frontier in the landscape of neural network research~\cite{maas1997networks,zenke2021visualizing}.

Similar to other types of neural networks, SNNs are now facing the problem of vulnerability to adversarial attacks.
In safety-critical areas where system reliability is crucial, low system reliability will hinder its widespread application, particularly in applications like autonomous driving~\cite{yamazaki2022spiking} and robotic control~\cite{bing2018survey}. Adversarial attacks, known for generating imperceptible perturbations that can mislead neural networks, pose a significant threat to the reliable functioning of neural networks~\cite{goodfellow2014explaining, szegedy2013intriguing,ozdenizci2021training}. Although some researchers found that special configurations of SNN can unleash their potential for greater robustness~\cite{sharmin2020inherent}, SNNs' susceptibility to adversarial attacks is a recognized concern. More recent research highlights the vulnerability of SNNs to adversarial attacks~\cite{kundu2021hire,alberto2021dvs,ding2022snn,bu2023rate,hao2024threaten}, underscoring the necessity to understand and improve their robustness. Currently, research focuses on how to leverage adversarial defense to improve the robustness of SNNs.

Deep SNNs usually use the leaky integrate-and-fire (LIF) neuron model. The dynamics of the LIF neuron consist of a leaky factor that controls the preserved information in the membrane potential. In the context of robustness, Sharmin et al.~\yrcite{sharmin2020inherent} found that the leak factor offers an additional control to manipulate adversarial perturbation. Different leaky factors correspond to different levels of smoothness in the noise. Further evidence of its importance can be revealed from the work of El-Allami et al.~\yrcite{el2021securing}. To improve the robustness, they manually traversed the leaky factor within some ranges and successfully found a robust configuration. Both works emphasize the effectiveness of properly setting the leaky factor, which results in different neuronal dynamics and thus emphasizes the importance of neuronal dynamics. On the other hand, other previous works have only focused on changes in discrete spike output. Instead of directly constraining discrete outputs, which may potentially lead to the problem of inaccurate supervision signals, these work shifted the focus to weight constraints or the use of adversarial training~\cite{kundu2021hire,ding2022snn,liang2022toward}. Therefore, we want to return to the dynamic nature of SNNs to study and find ways to resist perturbations.



Since the dynamics of SNN can help reduce the impact of noise, one question arises: how can we design beneficial dynamics to improve the robustness of SNN? An intuitive solution is to design a dynamic system with stability, which involves employing strategies that ensure the system's behavior remains bounded and converges to a desired state over time~\cite{khalil2002nonlinear}. SNN can be viewed as a learnable nonlinear dynamic system; therefore, we can also adopt wisdom from nonlinear dynamics and analyze SNN in similar ways.
This article aims to study the impact of perturbation on spiking neural networks from the perspective of nonlinear system stability and propose methods to improve the robustness of spiking neural networks. Our contribution can be summarized as follows:

\begin{itemize}
    \item Based on the dynamic equations of LIF neurons before and after perturbation, we obtain the membrane potential perturbation dynamics. Compared to discrete spike-based metrics, simplified membrane potential perturbation dynamics can serve as a reliable estimate of the impact of input perturbations on neuronal dynamics.
    \item We propose to improve the robustness by reducing the mean square of the membrane potential perturbation. In addition, our theoretical analyses prove that the membrane potential perturbation dynamics satisfy $L_2$ input-output stability. 
    \item We propose a training framework to improve the robustness of SNN by reducing the mean square of the membrane potential perturbation for the last neuron layer. Moreover, to further improve the reduction efficiency of $L_2$ gain, a dynamic LIF neuron is proposed to replace LIF neurons in SNN. 
    \item Our experiments show the effectiveness of the overall training framework, which significantly improves adversarial robustness in image recognition on the CIFAR-10 and CIFAR-100 datasets.
\end{itemize}

\section{Background and Related Work}


\subsection{Spiking Neural Networks}
\label{sec:snn}

Spiking neural networks emulate the behavior of natural neurons by deploying differential equations evolving over time. One of the most used neuron models in deep learning is the leaky integrate-and-fire model (LIF) \cite{kim2021revisiting, gerstner2014neuronal,xu2022hierarchical,shi2024towards}. The discrete form of the differential equation of LIF neurons in a deep SNN can be expressed as follows:
\begin{align}
    v_{i}^{l}\left[ t \right] &=\lambda u_{i}^{l}\left[ t-1 \right] +\sum_j{w_{ij}^{l}s_{j}^{l-1}\left[ t \right]}, \nonumber
\\
s_{i}^{l}\left[ t \right] &=H\left( v_{i}^{l}\left[ t \right] -u_{th} \right) , \label{eq:lif}
\\
u_{i}^{l}\left[ t \right] &=v_{i}^{l}\left[ t \right] \left( 1-s_{i}^{l}\left[ t \right] \right) . \nonumber
\end{align}
Here, $v_{i}^{l}[t]$ denotes the membrane potential of the $i$-th neuron in layer $l$ at time-step $t$ ($t=1,2,\cdots,T; l=1,2,\cdots,L; u_{i}^{l}[0 ]=0$), $s_{i}^{l}$ is the corresponding binarized spike generated when $v_{i}^{l}[t]$ crosses the threshold $u_{th}$ ($H$ is the Heaviside function). The membrane potential after generating the spike ($u_{i}^{l}[t]$) returns to resting potential (0), waiting for decaying by leaky factor $\lambda$ and receiving weighted input spikes ($\sum_j{w_{ij}^{l}s_{j}^{l-1}}$) from neurons in the preceding layer.

\subsection{Adversarial Attacks}

Neural networks are notorious for being able to be fooled by subtle perturbations in input data called adversarial attacks. The sad situation also holds for SNN, which has a higher sparsity of activation than ANN. One prevalent attack method is to express the attack by maximizing network loss $L$ such that a classifier $h:\mathcal{R}^d \rightarrow Y$, where $Y$ is the space for labels. Receiving input $x$ with perturbation $\boldsymbol{\delta}$ will result in misclassification $h(\boldsymbol{x}+\boldsymbol{\delta}) \neq h(\boldsymbol{x})$. The attacks should be imperceptible by applying guarantees that $\Vert \boldsymbol{\delta} \Vert_p \leq \epsilon$, where $\epsilon$ is typically an integer multiple of $1/255$ for images, and $p$ indicates the $p$-norm space. Formally, this optimization can be expressed as:
\begin{equation}
    \boldsymbol{\delta} =\underset{\Vert \boldsymbol{\delta} \Vert_p \leq \epsilon }{\text{arg}\max} \mathcal{L} \left( h(\boldsymbol{x} + \boldsymbol{\delta }),y \right).
\end{equation}
We denote the perturbed input as $\tilde{\boldsymbol{x}} = \boldsymbol{x} + \boldsymbol{\delta}$ for simplicity, and the superscript of tilde is used over the hidden variables related to $\tilde{\boldsymbol{x}}$ in the following content.

FGSM, introduced by Goodfellow et al.~\yrcite{goodfellow2014explaining}, is a fundamental attack method for creating adversarial examples by perturbing data in the negative direction of the gradient sign.
The following formula sums up this idea:
\begin{equation}
\boldsymbol{\tilde{x}}=\boldsymbol{x}+\epsilon~ \mathrm{sgn}\left( \nabla _{\boldsymbol{x}} \mathcal{L} \left( h(\boldsymbol{x}),y \right) \right) .
\end{equation}
Madry et al.~\yrcite{aleksander2018towards} propose an iterative version of FGSM, known as PGD, which is an efficient attack method that improves perturbations iteratively. It can be expressed as:
\begin{equation}
\boldsymbol{\tilde{x}}_{k+1}=\prod\nolimits_{\epsilon}^{}{\left( \boldsymbol{\tilde{x}}_k+\pi ~\mathrm{sgn} \left( \nabla _{\boldsymbol{x}} \mathcal{L} \left( h(\boldsymbol{\tilde{x}}_k),y \right) \right) \right)},
\end{equation}
where $\prod\nolimits_{\epsilon}$ ensures that the poisoned data is confined within the $p$-norm space around the clean data $\boldsymbol{x}$ and $\pi$ is the step size of one PGD iteration.
Previous work highlighted that SNN is vulnerable to the aforementioned crafted perturbations in certain input coding schemes like constant input coding~\cite{kundu2021hire}, while some suggested that using stochastic coding alternatives such as rate coding~\cite{sharmin2020inherent,sharmin2019comprehensive,ding2024enhancing} can enhance the security of SNN. Yet, the vulnerabilities can also be exposed by deploying surrogate functions for the Heaviside function and obtaining an SNN-specified attack with diverse attack methodologies. This essentially poses a threat to the wide deployment of neuromorphic hardware in safety-critical applications~\cite{liang2021exploring, liang2022toward}.

\subsection{Defensive Tools for SNN}

Borrowing wisdom from adversarial robustness on ANN, one can effectively build up a defensive network by exploiting adversarial training against attacks~\citep{kurakin2016adversarial, zhang2019theoretically}. This involves pushing deep networks to generalize on adversarial examples. While empirically effective, adversarial training faces limitations of generalization on unseen attacks~\citep{aleksander2018towards}, leaving room for potential vulnerabilities after training.

Researchers on SNN have worked to improve resilience by deploying unique techniques. Kundu et al.~\yrcite{kundu2021hire} improved the robustness of SNN by perturbing the images over time and performing adversarial training. Furthermore, recent advances have resulted in a larger improvement in robustness. Ding et al.~\yrcite{ding2022snn} developed regularized adversarial training (RAT) from the perspective of Lipschitz analysis. Meanwhile, methods proposed by Liang et al.~\yrcite{liang2022toward} explored the application of certified robustness on SNN by explicitly sensing the boundary of spike nonlinearity. These two works explore the temporal characteristics composed of machine learning tools, which inspires us to find new mathematical principles to analyze their robustness. Other empirical findings on enhancing the robustness of SNN emphasized the importance of both novel training methodologies and structural optimizations in advancing the security of these models~\citep{el2021securing, sharmin2019comprehensive}.

\subsection{Input-Output Stability}

SNNs in Eqs.~\ref{eq:lif} can be considered as a nonlinear input-output system. We would like to note that it is quite common to view a neural network as a complex dynamic system. Previous works are mostly on temporally continuous ANNs that target robotic control, physical systems, and biological systems~\cite{kojima2022learning,lawrence2020almost,chen2018neural}. The challenge of SNN lies in the unknown guarantee that SNN can have after training. By examining the second row in Eqs.~\ref{eq:lif}, one can get the impression that the output of the spiking neuron is bounded, which is supposed to be capable of tolerating more input noise. Liang et al.~\yrcite{liang2022toward} gave a linear relaxation of the sparse Heaviside function and formulated the input boundaries of spike inputs. This paper will give a theoretical point of view on the stability of SNN under perturbations. 

We introduce the $L_2$ input-output stability here~\cite{khalil2002nonlinear}. $L_2$ stability measures the ability of a system to maintain boundedness in the norm ratio between the output signal and the input signal. The norm ratio, called $L_2$ gain, quantifies the stability of a system. The $L_2$ norm is employed in the definition of $L_2$ stability and is calculated over the spaces of input and output signals. Specifically, consider a system $\boldsymbol y = f ( \boldsymbol x),$
where $f$ is some operator that relates $\boldsymbol y$ and $\boldsymbol x$. $\boldsymbol x$ has a temporal axis in $\left[ 0, \infty\right)$ and is defined in Euclidean space $R^m$. The $L_2$ norm of a signal $\boldsymbol x$ is given by the expression $\Vert \boldsymbol x \Vert_{L_2}=\sqrt{ \int^\infty_0 \Vert \boldsymbol x(t) \Vert^2dt}$, which provides a quantitative measure of the signal's energy. To address $L_2$ stability in the context of the nonlinear system, typically an assumption regarding the origin $\boldsymbol x \equiv 0$ of the nonlinear system is that this origin is an asymptotically stable equilibrium point and gives $f(0)=0$. Definition~\ref{def:l2_stable} outlines the $L_2$ norm and introduces the criteria for $L_2$ stability.

\begin{definition}
    \label{def:l2_stable}
    \cite{khalil2002nonlinear}~For a nonlinear system $\boldsymbol y = f ( \boldsymbol x)$, the $L_2$ norm of signal $\boldsymbol x$ is $\Vert \boldsymbol x \Vert_{L_2}=\sqrt{ \int^\infty_0 \Vert \boldsymbol x(t) \Vert^2dt}$. $\boldsymbol x_{[:\tau]}(t)$ denotes signal $\boldsymbol x(t) , (0\leq t \leq \tau)$ . If there exists a continuous function $\alpha:\left[0,\infty\right) \rightarrow \left[0,\infty\right)$ belonging to class $\kappa$ and a non-negative constant $\beta$, such that for all $\boldsymbol x$ and $\tau \in \left[ 0,\infty \right)$,
\begin{equation}
    \Vert f(\boldsymbol x)_{[:\tau]} \Vert_{L_2} \leq \alpha ( \Vert \boldsymbol x_{[:\tau]} \Vert_{L_2} ) + \beta.
\end{equation}
Then, the system is $L_2$ stable. If there exist non-negative constants $\gamma$ and $\beta$, such that for all $\boldsymbol x$ and $\tau \in \left[ 0,\infty \right)$,
\begin{equation}
\Vert f(\boldsymbol x)_{[:\tau]} \Vert_{L_2} \leq \gamma \Vert \boldsymbol x_{[:\tau]} \Vert_{L_2} + \beta.
\end{equation}
Then, the system is finite-gain $L_2$ stable, where the minimum $\gamma$ is called the $L_2$ gain of the system.
\end{definition}

\section{Stable Spiking Neural Networks}

In this section, we will analyze the neuronal dynamics under attack and derive modified dynamics for input perturbations. We call it the membrane potential perturbation dynamics. This dynamics can be referred to as an accurate indicator of how much the network is perturbed. By designing and minimizing the mean square of the potential perturbation, we can improve the robustness of SNNs.

\subsection{Neuronal Dynamics for Input Perturbations}

\begin{figure*}[t]
    \centering
\setcounter{subfigure}{0}
    \subfigure[Membrane potential perturbation (MPP) dynamics]{
\includegraphics[width=0.32\linewidth]{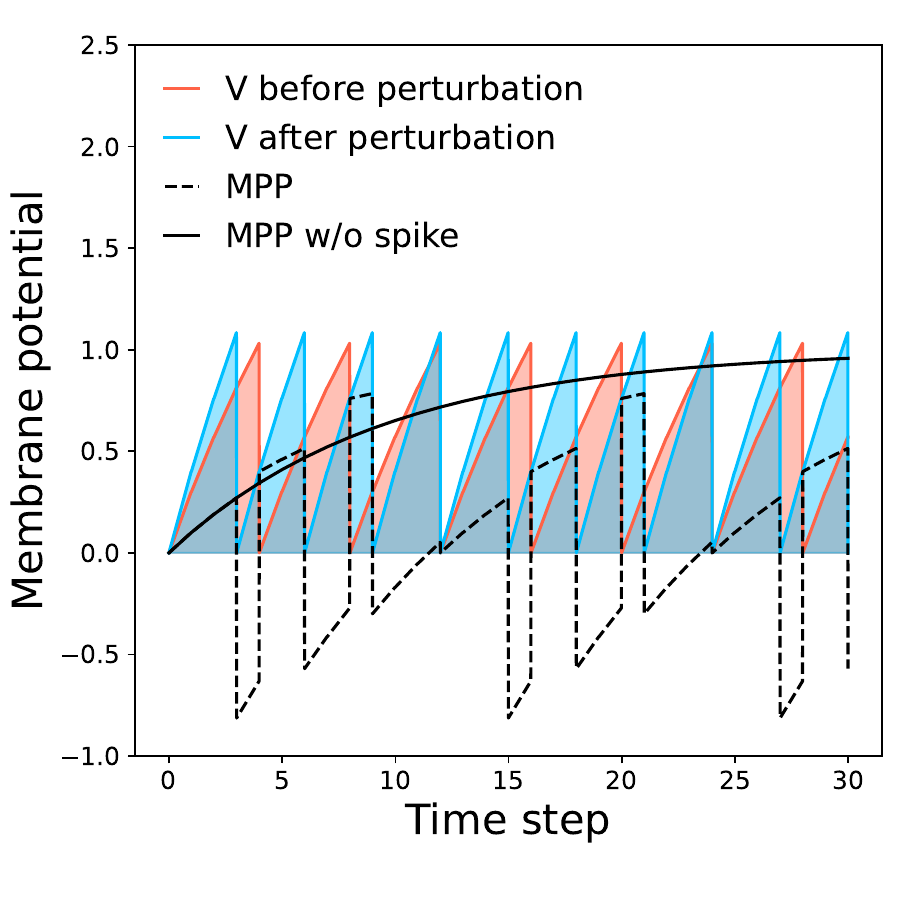}}
\subfigure[Constant perturbation]{
\includegraphics[width=0.32\linewidth]{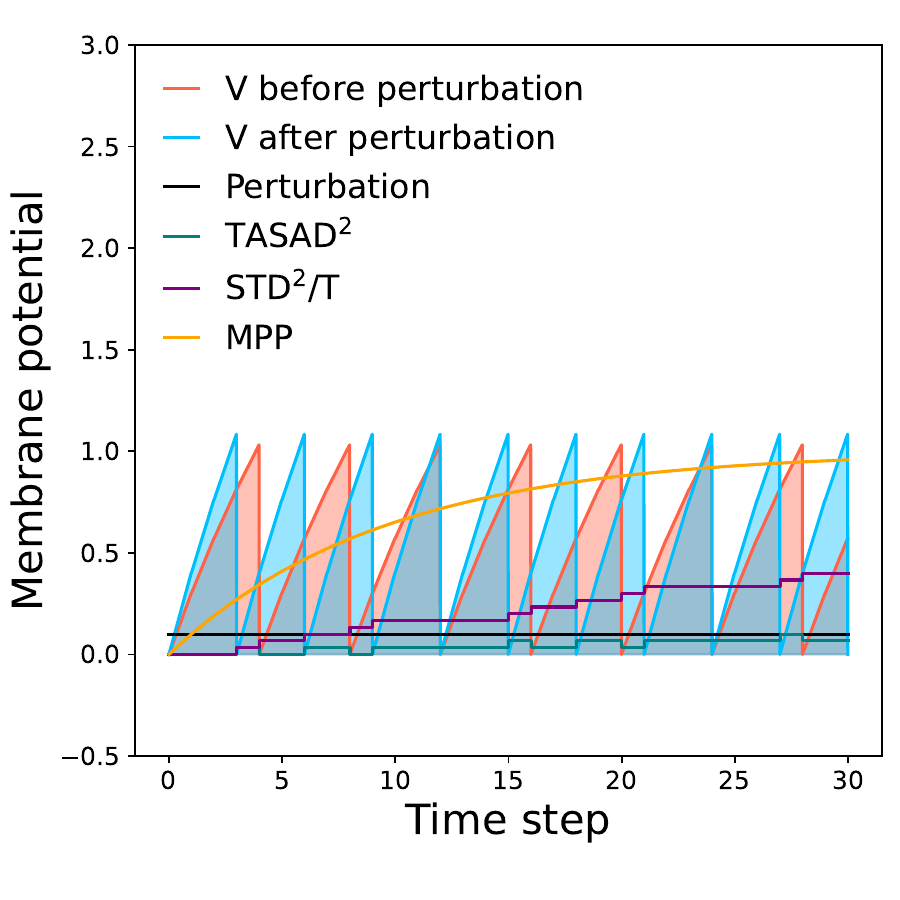}}
\subfigure[Temporal Gaussian noise]{
\includegraphics[width=0.32\linewidth]{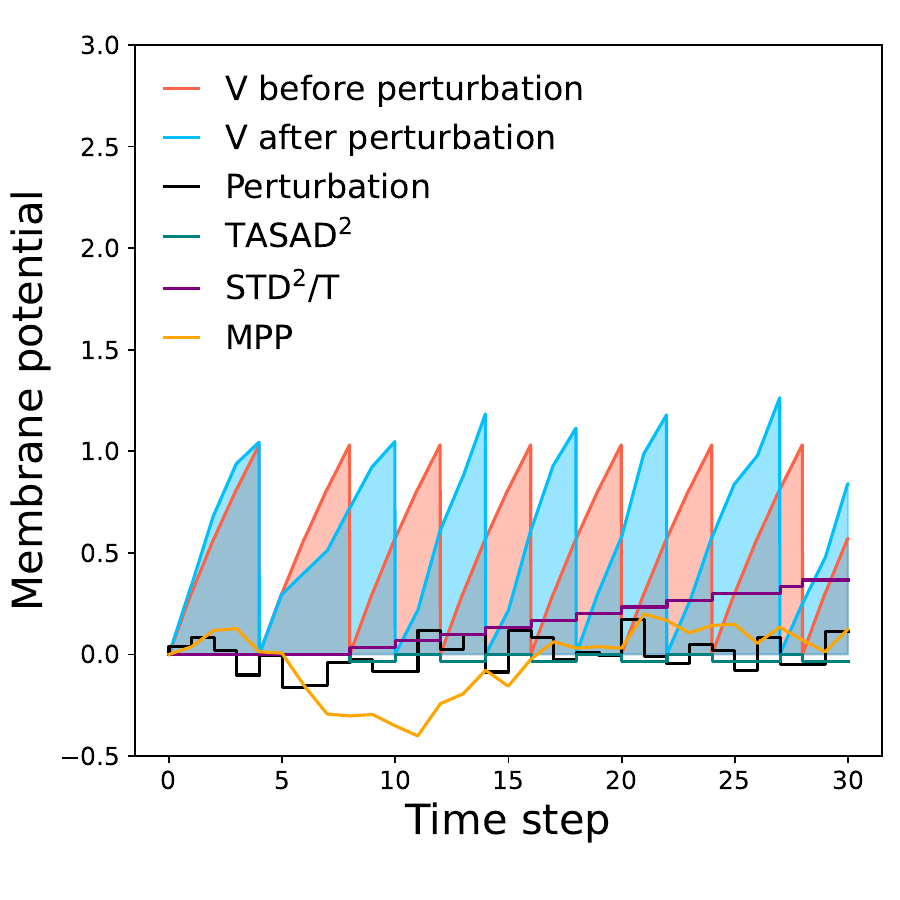}}
    \caption{Illustration of the membrane potential perturbation (MPP) dynamics. The LIF neuron in all subfigures receives a constant input of 0.3$u_{th}$. In (a)(b), the perturbation is +0.1$u_{th}$. In (c), the perturbation is sampled from a Gaussian distribution $\mathcal N(0, (0.3u_{th})^2)$. }
    \label{fig:perturbation_dynamic}
\end{figure*}

By recording the float-point internal variables of spiking neurons in Eqs.~\ref{eq:lif}, if the input sequence is under perturbation, we can get some knowledge of the noise budget forced to SNN. For a hidden layer $l$, we denote $\tilde{s}_{j}^{l-1}$ as the perturbed spike sequence from neuron $j$ in the presynaptic layer $l-1$. Let first investigate the initial stage of one neuron before its first spike: $s_{i}^{l}\left[ t_0 \right] =1$, that is, $1 \leq t \leq t_0$. By subtracting the clean and perturbed version of the first equation in Eqs.~\ref{eq:lif}, we can derive the dynamics with regard to the perturbation:
\begin{align}
\left( v_{i}^{l}\left[ t \right] -\tilde{v}_{i}^{l}\left[ t \right] \right) =\lambda \left( v_{i}^{l}\left[ t-1 \right] -\tilde{v}_{i}^{l}\left[ t-1 \right] \right) + \nonumber \\ 
\sum_j{w_{ij}^{l}\left( s_{j}^{l-1}\left[ t \right] -\tilde{s}_{j}^{l-1}\left[ t \right] \right)},~1 \leq t \leq t_0, \label{eq:membrane_difference}
\end{align}
where the denotation with the superscript tilde is the perturbed version of the origin variable. When $t=0$, $v_{i}^{l}$ and $\tilde{v}_{i}^{l}$ equal zero. If we denote the difference of the membrane potential $v_{i}^{l}\left[ t \right]$ before and after perturbation as $\varepsilon _{i}^{l}\left[ t \right] =v_{i}^{l}\left[ t \right] -\tilde{v}_{i}^{l}\left[ t \right] $ ($\varepsilon _{i}^{l}[0 ]=0$) and the difference of presynaptic spike train as $\varDelta s_{j}^{l-1}\left[ t \right] =s_{j}^{l-1}\left[ t \right] -\tilde{s}_{j}^{l-1}\left[ t \right] $, then we can simplify Eq.~\ref{eq:membrane_difference} to:
\begin{equation}
\varepsilon _{i}^{l}\left[ t \right] =\lambda \varepsilon _{i}^{l}\left[ t-1 \right] +\sum_j{w_{ij}^{l}\varDelta s_{j}^{l-1}\left[ t \right]},~1 \leq t \leq t_0. \label{eq:membrane_difference_simple}
\end{equation}
Eq.~\ref{eq:membrane_difference_simple} is actually an iterative equation with regard to the membrane potential perturbation. The dynamics of $\varepsilon$ strictly characterize the change in membrane potential affected by perturbations. One limitation of this equation is that it will not hold after the neuron fires a spike. When the limit of $1 \leq t \leq t_0$ is removed, we assume $\varepsilon _{i}^{l}\left[ t \right] =v_{i}^{l}\left[ t \right] -\tilde{v}_{i}^{l}\left[ t \right] $ for all time steps and can have:
\begin{align}
\varepsilon _{i}^{l}\left[ t \right] &=\lambda \varepsilon _{i}^{l}\left[ t-1 \right] +J, \label{eq:J}
\\ 
J&=\sum_j{w_{ij}^{l}\varDelta s_{j}^{l-1}\left[ t \right]} \nonumber \\
-\lambda &\left( v_{i}^{l}\left[ t-1 \right] s_{i}^{l}\left[ t-1 \right] -\tilde{v}_{i}^{l}\left[ t-1 \right] \tilde{s}_{i}^{l}\left[ t-1 \right] \right) .\label{eq:J1}
\end{align}
Here in Eq.~\ref{eq:J}, $J$ denotes the input for the dynamics of $\varepsilon$. We can observe that in addition to considering the influence of the weighted sum caused by the perturbation, this input is also affected by the neuronal reset of the neuron. If there is a spike in the previous time step, the dynamics of $\varepsilon$ fluctuate due to the neuronal reset. To get the dynamics of $\varepsilon$ rid of the fluctuation, we propose to reduce the resetting part in Eq.~\ref{eq:J1}. Eqs.~\ref{eq:J} and \ref{eq:J1} together construct a dynamics, which we name membrane potential perturbation dynamics (MPPD). For layer $l$ in SNN running for $T$ time steps,
\begin{align}
\textit{MPPD}:~\varepsilon _{i}^{l}\left[ t \right] =\lambda \varepsilon _{i}^{l}\left[ t-1 \right] +\sum_j{w_{ij}^{l}\varDelta s_{j}^{l-1}\left[ t \right]},\\ 
t=1,2,\cdots ,T. \nonumber
\end{align}
Figure~\ref{fig:perturbation_dynamic}(a) shows the difference between the simplified perturbation dynamics and that before simplification. We input constant currents of 0.3$u_{th}$ (before perturbation, red line) and 0.4$u_{th}$ (after perturbation, blue line) to the LIF neuron. The difference in membrane potentials changing with time in these two cases is the unsimplified perturbation dynamics (dotted black line). The dotted line jitters violently with time steps. In contrast, the simplified perturbation dynamics have a very smooth curve (solid black line) since there is no reset effect. Its evolution can reflect the leaky factor $\lambda$ of neuronal dynamics.

\subsection{Metric for Measuring Perturbation}
\label{sec:sensitivity}

Previous studies on the robustness of SNN often involved a proposal of distance under perturbation as a measurement of how much the neuronal dynamics is affected by the perturbation. From this point of view, the proposed perturbation dynamics can inherently be recognized as a metric measuring the perturbation. We are going to compare the proposed metric to the time-averaged spiking activity distance (TASAD) proposed by Kundu et al.~\yrcite{kundu2021hire} and spike train distance (STD) proposed by Ding et al.~\yrcite{ding2022snn} in terms of the sensitivity to input noise.

TASAD, as implied by its name, calculates the distance between average firing rates before and after perturbation, which emphasizes the patterns of firing rate. Using the notation system in Section~\ref{sec:snn}, TASAD can be expressed as:
\begin{equation}
\textit{TASAD}^l=\left\| \left( \sum_{t=1}^T{\boldsymbol{s}^l\left[ t \right]}-\sum_{t=1}^T{\boldsymbol{\tilde{s}}^l\left[ t \right]} \right) /T \right\| _2,
\end{equation}
As items in $\boldsymbol{s}^l\left[ t \right]$ only take values in 0 and 1, TASAD operates in a discrete space and quantifies changes in spiking activity over a period. It works especially well for rate-coded SNNs.

STD, different from TASAD, quantifies the difference in spike counts before and after perturbation. It provides a metric for evaluating the impact on the overall spiking pattern. STD can be expressed as:
\begin{equation}
\textit{STD}^l=\sqrt{\sum_{t=1}^T{\left\| \boldsymbol{s}^l\left[ t \right] -\boldsymbol{\tilde{s}}^l\left[ t \right] \right\| _{2}^{2}}}, \label{eq:std}
\end{equation}
According to Eq.~\ref{eq:std}, even without changes in firing rate, STD can be non-zero and sensitive to variations in spike counts due to changes in spike time. Similar to TASAD, STD is also constrained to discrete spaces due to its reliance on spike counts.

In contrast, our proposed membrane potential perturbation introduces a unique approach by considering float-point input differences and applying neuronal dynamics without spiking operations. This approach allows the potential perturbation to operate in continuous spaces, providing a more fine-grained sensitivity to input noise levels. Suppose $J$ in $\varepsilon _{i}^{l}\left[ t \right] =\lambda \varepsilon _{i}^{l}\left[ t-1 \right] +J$ is a constant perturbation and $\varepsilon _{i}^{l}\left[ 0 \right]=0$, then $\varepsilon _{i}^{l}\left[ t \right] =\frac{1-\lambda ^t}{1-\lambda}J$ faithfully reflect the effect of leaky factor and noisy input. We presume $J$ to be constant because our SNNs, similar to the approach outlined by Kim et al.~\yrcite{kim2022rate}, employ direct coding where constant images are fed into the first layer. Consequently, J remains constant within this layer.

We simulate a LIF neuron for 30 time steps and record the membrane potential perturbation before and after perturbation. In terms of perturbation type, we choose to add a constant bias input current and to add a temporal Gaussian noise to the neuron. The results are illustrated in Fig.~\ref{fig:perturbation_dynamic}(b)(c). The quantization effect of STD and TASAD can be clearly seen in the figures. When we add the constant perturbation to the input current, the TASAD curve fluctuates severely due to the irregular spike occurrence. STD and MPPD can indicate the intensity of the added perturbation. When we add temporal Gaussian noise to the input current, STD can only signify the existence of the noise but not the temporal change of the noise. In comparison, the MPPD curve shows a smoothed version of Gaussian noise. Note that temporal Gaussian noise can be evoked in rate-coding SNN due to the irregular spike occurrence weighted by Gaussian-distributed synaptic weights. 

Perturbation dynamics are more sensitive to input noise, which means that reducing a specific moment estimator of membrane potential perturbation can lead to better robustness. Therefore, we propose to minimize the mean square of MPPD (MS-MPPD) for the last neuron layer (layer $L$) to align the features between perturbed input and clean input:
\begin{equation}
    \textit{MS-MPPD}^L=
    \sum\nolimits_{i=1}^{N^L}{\sum\nolimits_{t=1}^T{\left( \varepsilon _{i}^{L}\left[ t \right] \right) ^2}},
\end{equation}
where $N^L$ is the number of neurons in layer $L$, $T$ is the number of time steps.


\begin{figure*}[t]
    \centering
    \includegraphics[width=0.9\textwidth]{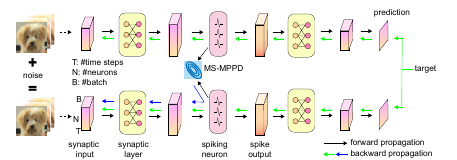}
    \caption{Training paradigm of our robust stable SNN.}
    \label{fig:training}
\end{figure*}

\subsection{Determining the Stability}

When training weights to reduce MS-MPPD, we limit the extent to which noise affects features.
Here we would like to derive the property of $L_2$ stability for the perturbation dynamics.

\begin{theorem}
\label{thm:stability}
    Given the membrane potential perturbation dynamics of SNN inferring for $T$ time steps as $\boldsymbol{\varepsilon }^l\left[ t \right] =\lambda \boldsymbol{\varepsilon }^l\left[ t-1 \right] +\boldsymbol{W}^l\varDelta \boldsymbol{s}^{l-1}\left[ t \right] $ for layer $l$, where $\boldsymbol{W}^l$ is the weight matrix of layer $l$, $\varDelta \boldsymbol{s}^{l-1}$ is the perturbation from layer $l-1$, $\boldsymbol{\varepsilon }^l\left[ 0 \right]=\boldsymbol{0}$, we have
    \begin{equation}
        \left\| {\boldsymbol{\varepsilon }^l}_{[:T ]} \right\| _{L_2}\leqslant \gamma ^l\left\| \varDelta {\boldsymbol{s}^{l-1}}_{[:T ]} \right\| _{L_2}+\beta ^l,
    \end{equation}
    where $\gamma^l=\sqrt{1/(1-\lambda)} \Vert \boldsymbol{W}^l \Vert$ and $\beta^l=0$. $\Vert \boldsymbol{W}^l \Vert$ is the spectral norm of the weight.
\end{theorem}

From Definition~\ref{def:l2_stable}, $\gamma^l$ is the $L_2$ gain of the perturbation dynamics. Theorem~\ref{thm:stability} suggests a promoting mechanism to maximize the capability of controlling the $L_2$ gain. 
For the detailed derivation process of Theorem~\ref{thm:stability}, please refer to the Appendix. Yet, $\gamma^l$ here is not a tight bound for the $L_2$ gain. 
$\sqrt{1/(1-\lambda)}$ in $\gamma^l$ reflects the effect of time-step iteration, which can be found in the proof. This effect finally manifests itself as the summation of powers of leaky factors with an upper bound of $\sqrt{1/(1-\lambda)}$.

Inspired by the work of PLIF with trainable leaky time constant~\cite{fang2021incorporating}, we propose to add a trainable dynamic parameter $a^l[t]$ to neurons for input of each time step such that $v_{i}^{l}\left[ t \right] =\lambda a^l[t] u_{i}^{l}\left[ t-1 \right] +\sum_j{w_{ij}^{l}s_{j}^{l-1}\left[ t \right]}$. We call this altered type of LIF neuron here Dynamic LIF (DLIF). The added $a^l[t]$ can have an individual effect on minimizing $\gamma^l$ orthogonal to the effect of altering $\boldsymbol{W}^l$. Therefore, we can replace the original LIF neuron in SNN with DLIF and attempt to minimize MS-MPPD while training.

\begin{table*}[t]
\caption{Performance of our robust stable SNN compared with current state-of-the-art work. }
\label{tab:sota}
\centering
\small
\setlength\tabcolsep{4pt}
\begin{tabular}{lllllllll}
\toprule
Model               & Clean & FGSM & PGD$^7$ & PGD$^{10}$ & PGD$^{20}$ & PGD$^{40}$ & APGD$^{10}_{\text{CE}}$ & APGD$^{10}_{\text{DLR}}$ \\ \midrule
\multicolumn{9}{c}{CIFAR10}                                                           \\ \midrule
SNN-BP, VGG5~\yrcite{sharmin2020inherent}         & 89.3  &  15.0 &  3.8  &     -  &   -    &    -   &     -      &    -        \\
HIRE-SNN, VGG5~\yrcite{kundu2021hire}       &  87.9  &  35.5  &  5.3    &   -    &    -   &     -  &    -       &     -       \\
SNN-RAT, VGG11~\yrcite{ding2022snn}        & 90.74  & 45.23 & 21.16 &  -    &  -     &    -   &      -     & -           \\
LIF, VGG11, Natural & 92.54   &  10.33    &  0.03 & 0.01  & 0.01  & 0.01   & 0.02     & 0.05    \\
DLIF, VGG11, Natural & 92.22  &    13.24  & 0.09  & 0.02  & 0.01  &  0.01  &    0.05  &  0.03   \\

DLIF, VGG11, Gaussian, $\rho=0.0$  &  92.43 &   11.30   & 0.18  & 0.10  &  0.08  &  0.06 &   0.08   &   0.09  \\
DLIF, VGG11, Gaussian, $\rho=1.0$  &   92.39  &  15.24    &    0.23  &  0.09   &  0.08 &  0.10    &     0.17   &  0.08   \\
DLIF, VGG11, AT, $\rho=0.0$  &  90.07   &  43.54    &   30.57   &  29.06   & 28.53  &   28.00   &    23.05    &  29.88   \\
DLIF, VGG11, AT, $\rho=1.0$  &   87.21  &  49.02    &  38.68    &  37.55   &  37.08 &    36.41  &    33.25    &  39.68   \\

DLIF, VGG11, AT+Reg, $\rho=0.0$  &  89.61   &  52.10    & 34.83     &  32.01  &  29.98 &   28.63   &    29.07   &  33.67   \\
DLIF, VGG11, AT+Reg, $\rho=1.0$  &  88.91   &    56.71  &   40.30   &  37.53   &  35.25  &   33.93   &  35.09     &   39.85 \\

LIF, WRN16, Natural & 94.28  &  12.80    &  0.00 &  0.00 & 0.00  & 0.00   &   0.00   &  0.00   \\
DLIF, WRN16, Natural &  94.01 &   12.29   &  0.00 &  0.00 & 0.00  &  0.00  &    0.00  &  0.00   \\
DLIF, WRN16, Gaussian, $\rho=0.0$       &  93.88    &  11.41 & 0.03  &  0.02 & 0.00 & 0.00   &    0.01  &  0.02       \\
DLIF, WRN16, Gaussian, $\rho=1.0$  &  92.85   &   13.40   &   0.07   &   0.06  & 0.05  &  0.03    &    0.03    &  0.08   \\
DLIF, WRN16, AT, $\rho=0.0$   &   90.16  &    48.09  &   33.70   &  32.37  & 31.33  &    30.99  &    27.94    &  32.39   \\
DLIF, WRN16, AT, $\rho=1.0$   &   90.11  &   49.82   &   36.21   &    34.71 & 33.71  &    33.41  &   30.08     &  34.85   \\
DLIF, WRN16, AT+Reg, $\rho=0.0$  &   91.38  &    56.87  &  36.77    &  33.24   & 30.49  &  29.07    &     31.38   &   34.55  \\
DLIF, WRN16, AT+Reg, $\rho=1.0$  &   91.15  &  57.89    &   38.78   &  35.33   &  32.82  & 31.13      &  32.90      &  37.39   \\
\midrule

\multicolumn{9}{c}{CIFAR100}                                                           \\ \bottomrule
SNN-BP, VGG11~\yrcite{sharmin2020inherent}         & 64.4 & 15.5 & 6.3 &    -   &    -   &     -  &     -      &     -       \\
HIRE-SNN, VGG11~\yrcite{kundu2021hire}       &     65.6 & 16.4 & 2.9  &  -&   -    &    -   &    -       &  -          \\
SNN-RAT~\yrcite{ding2022snn}        & 70.89  & 25.86  &  17.81  &    -   &    -   &  -     &      -     &    -       \\
LIF, VGG11, Natural &  72.48 &   5.33   &  0.06  & 0.03  & 0.01  &  0.02  &   0.03   & 0.14    \\
DLIF, VGG11, Natural & 70.79  &   6.95   &  0.08 &  0.05 &  0.00 &  0.00  &    0.02  &  0.18   \\

DLIF, VGG11, Gaussian, $\rho=0.0$     & 70.82  & 7.99 & 0.56  & 0.51 & 0.39 &   0.38 &  0.33  &  0.86 \\
DLIF, VGG11, Gaussian, $\rho=1.0$  &   70.51  &   8.72   &   0.77   &   0.55  & 0.52  &   0.47   &    0.48    &  0.94   \\
DLIF, VGG11, AT, $\rho=0.0$  & 63.35    &  24.97   &  16.61   &   15.99  &  15.49 &   15.32  &       13.64  &  16.81   \\
DLIF, VGG11, AT, $\rho=1.0$  &   63.85  &  25.13    &  16.80    &   16.02  & 15.56  &    15.52  &     13.15   &   17.31  \\
DLIF, VGG11, AT+Reg, $\rho=0.0$  &  65.94   &   36.00   &  23.53   &   20.98  &  18.73 & 17.45     &     19.77   &   24.25  \\
DLIF, VGG11, AT+Reg, $\rho=1.0$  &  66.33   &    36.83  &   24.25   & 21.64    & 19.22  &  17.84    &    20.68    &   24.21  \\

LIF, WRN16, Natural & 73.06  &   7.49   & 0.00  &  0.00 & 0.00  & 0.00   &   0.00   &  0.11   \\
DLIF, WRN16, Natural &  73.85 & 8.08     &   0.00 &   0.00 &   0.00 &   0.00  &   0.00   &   0.09  \\
DLIF, WRN16, Gaussian, $\rho=0.0$  & 72.19  &  9.18    &  0.41 &  0.25  &   0.17  &  0.15  &   0.16   &   0.74  \\
DLIF, WRN16, Gaussian, $\rho=1.0$  & 68.87    &   9.10   &   0.65   &  0.44   & 0.36  &   0.34   &     0.38   &  1.22   \\
DLIF, WRN16, AT, $\rho=0.0$   &   65.86  &  25.90    &   15.20   &  14.03   & 13.37  &   13.30   &     11.98   &   16.32  \\
DLIF, WRN16, AT, $\rho=1.0$  &  65.26   &   25.73   &   16.22   &   15.11  & 14.68  &   14.09   &   13.09     &  16.93   \\
DLIF, WRN16, AT+Reg, $\rho=0.0$  &  66.57   &   33.05   &   18.75   &  16.23  & 14.16  &   13.44   &    14.93    &  20.53   \\
DLIF, WRN16, AT+Reg, $\rho=1.0$  &   65.58  &   33.56   &   19.22   &  17.14   & 15.52  &   14.41   &   15.87     &   21.68  \\
\bottomrule
\end{tabular}
\end{table*}

\subsection{Stabilizing Spiking Neural Networks}

According to the analysis in the subsection above, we propose a training framework to stabilize SNN against perturbation. The idea is to replace the traditional LIF neuron with a DLIF neuron and minimize MS-MPPD for the last spiking neuron layer in SNN. The training paradigm of our robust stable SNN is shown in Figure~\ref{fig:training}. 

Take the task of image classification for example. The framework first performs an adversarial attack or some type of perturbation on the original input, which will give an adversarial version of the input images. Then, the two versions of input will both be fed into SNN. The inputs of the last spiking layer will be recorded for clean and perturbed examples, respectively. Then, we subtract the two inputs and calculate $\textit{MS-MPPD}^L$. After that, the outputs of SNN corresponding to two inputs are used to obtain the task loss, combining the clean loss and the loss under perturbation. Following a mixup strategy~\cite{zhang2018mixup,wang2019improving}, the task loss can be expressed as:
\begin{equation}
    \mathcal{L}_{task} =\chi \mathrm{CE}\left( f_{SNN}\left( \boldsymbol{x} \right) ,y \right) +\left( 1-\chi \right) \mathrm{CE}\left( f_{SNN}\left( \boldsymbol{\tilde{x}} \right) ,y \right) ,
\end{equation}
where $\chi$ is a mixture parameter, which is 0.5 by default. 
Thus, the total loss of our proposed framework can be depicted as:
\begin{equation}
    \mathcal{L} =\mathcal{L} _{task}+\rho \cdot \textit{MS-MPPD}^L ,
    \label{eq:loss}
\end{equation}
where $\rho$ indicates the intensity of MS-MPPD. Minimizing MS-MPPD directly improves the similarity between the outputs corresponding to clean and perturbed inputs. The additional temporal parameter in DLIF will enhance the utility of maximizing similarity.

We use the STBP training algorithm to train SNNs. The core of STBP training is to enable backpropagation with surrogate functions instead of the non-differentiable Heaviside function. In this paper, we use the triangle-like surrogate functions~\cite{deng2021temporal}. It can be described as:
\begin{equation}
\frac{\partial s_{i}^{l}\left[ t \right]}{\partial v_{i}^{l}\left[ t \right]}=\frac{1}{\omega ^2}\max \left( \omega -\left| v_{i}^{l}\left[ t \right] -u_{th} \right|,0 \right) ,
\end{equation}
where $\omega=1$ by default. Note that the triangle-like surrogate function is also used to craft white-box adversarial examples in the proposed framework or robustness evaluation.

\begin{table*}[!t]
\caption{Ablation study.}
\label{tab:ablation_study}
\centering
\small
\setlength\tabcolsep{4pt}
\begin{tabular}{lllllllll}
\toprule
       Model       & Clean & FGSM & PGD$^7$ & PGD$^{10}$ & PGD$^{20}$ & PGD$^{40}$ & APGD$^{10}_{\text{CE}}$ & APGD$^{10}_{\text{DLR}}$ \\ \midrule
DLIF, AT, $\rho=0.0$, MS-MPPD &  85.32  &  38.61  &   27.28   &  26.27  &  25.78  &  25.72  &      22.37     &     26.43       \\
DLIF, AT, $\rho=0.5$, MS-MPPD &  85.01  &  38.82  &   27.47   &  26.36  &  25.88  &  25.83 &      22.80     &     26.30       \\ 
DLIF, AT, $\rho=1.0$, MS-MPPD &  85.21  &  39.63  &   \textbf{28.33}   &  \textbf{27.34}  &   \textbf{26.98}  &  \textbf{26.36} &     \textbf{23.94}      &      \textbf{27.70}    \\ 
DLIF, AT, $\rho=2.0$, MS-MPPD &  85.15  &  39.41  &   27.35   &  26.29  &  25.62  &  25.53  &      22.95     &     26.71     \\ \midrule
LIF, AT, $\rho=0.0$, MS-MPPD &  \textbf{85.61}   &  39.78  &  27.54  & 26.38  &  25.75   &  25.17  &   22.95    &     26.43    \\
LIF, AT, $\rho=0.5$, MS-MPPD &  85.11   & 39.19   & 27.29   &  26.20 &  25.73   & 25.48   &  22.81     &    26.75     \\
LIF, AT, $\rho=1.0$, MS-MPPD &  85.11  &  \textbf{40.15} &   27.92   &  26.81  &  26.00  &  25.87  &      23.30     &     27.05     \\
LIF, AT, $\rho=2.0$, MS-MPPD &  64.62   &   30.51 &  22.42  &  21.38  &   20.87  &  20.63  &  19.35     &    22.62     \\ \midrule
DLIF, AT, $\rho=1.0$, TASAD &  85.28  &  39.26 &  27.80  & 26.85 & 26.12   & 25.92  &      22.92  &      26.85  \\
DLIF, AT, $\rho=1.0$, STD &   85.46  & 38.96  &  26.82   &  25.75  &   25.38  & 25.03  &   22.07    &  25.94  \\
\bottomrule
\end{tabular}

\end{table*}

\section{Experiments}

\subsection{Experimental Setup}

We conduct experiments to verify our method to construct a robust, stable SNN for the image classification task. We employ the architecture setting from the current SOTA SNN robustness work~\cite{ding2022snn}, using SNN versions of VGG11 and WideResNet-16-4 (WRN16) for the CIFAR-10 and CIFAR-100 datasets. The time step to infer SNN is set to 8 by default. To verify the effectiveness of the proposed framework, the perturbation is chosen to be Gaussian noise and adversarial noise (AT for short). The intensity of Gaussian noise is $\epsilon=8/255$. And the construction of adversarial noise follows RFGSM methods~\cite{wong2019fast}, with an initial random step of $0.001$ and a fast-gradient-sign step with $\epsilon=4/255$. We also verify the compatibility of our framework with the regularizer (Reg for short) proposed by Ding et al.~\yrcite{ding2022snn} for SNN. For detailed training hyperparameters, please refer to the Appendix.

While testing the performance of robustness, we choose FGSM~\cite{goodfellow2014explaining}, PGD~\cite{aleksander2018towards} and Auto-PGD (APGD)~\cite{croce2020reliable} attacks to construct adversarial examples for evaluation. $\epsilon$ for evaluation is set to $8/255$. The steps of PGD vary from 7 to 40. For APGD, we use the 10-step APGD of the loss of cross-entropy (CE) and difference-of-logits-ratio (DLR).

\subsection{Results}

We compare the performance of SNNs trained by our framework with current SOTA work in Table~\ref{tab:sota}. We denote the setting of network training with clean data as `natural' in the table. When $\rho=1$, this means we are minimizing MS-MPPD while training in Eq.~\ref{eq:loss}. By comparing the performance of $\rho=1$ and $\rho=0$, we can know the effectiveness of minimizing MS-MPPD.

For both CIFAR-10 and CIFAR-100, SNNs with natural training are vulnerable to strong PGD or APGD attacks. For VGG11 on CIFAR-10 and CIFAR-100, SNN with DLIF outperforms SNN with vanilla LIF in most cases of attack. This implies that DLIF itself has the capability of improving robustness, though it is not significant. When training with Gaussian noise, the performance of DLIF improves more when $\rho=1$. For example, the improvement is 3.94\% for VGG11 and 1.99\% for WRN16 on the CIFAR-10 dataset.

The improvement in performance is more prominent when training with adversarial noise. For VGG11 with DLIF, training with $\rho=1$ improves the performance of PGD$^{10}$, APGD$^{10}_\text{CE}$, and APGD$^{10}_\text{DLR}$ from 29.06\%, 23.05\%, and 29.88\%, respectively, to 37.55\%, 33.25\%, and 39.68\%, respectively, compared with those when $\rho=0$. We think the improvement is due to minimizing MS-MPPD, which has enhanced the similarity of internal representation between the perturbed and clean data. With the assistance of adversarial noise, our performance is showing supreme robustness against HIRE-SNN~\cite{kundu2021hire}, which also gains robustness through adversarial training.

By integrating the previously proposed regularizer in SNN-RAT~\cite{ding2022snn} for SNN into the framework, our model produced by the framework gives the best overall performance. Our regularized model with $\rho=1$ gives PGD$^7$ accuracy of 49.02\% and 56.71\% for VGG11 and WRN16, respectively, on CIFAR-10, higher than 45.23\% of SNN-RAT. Similarly, our regularized model with $\rho=1$ gives PGD$^7$ accuracy of 36.83\% and 33.56\% for VGG11 and WRN16, respectively, on CIFAR-100, higher than 25.86\% of SNN-RAT. Thus, we believe that with our architecture, our model can further achieve robust performance for SNN.

\subsection{Effect of $\rho$}

Table~\ref{tab:ablation_study} studies the effect of $\rho$. $\rho$ determines the intensity to increase the similarity of representations. We conduct experiments on VGG-5 on the CIFAR-10 dataset. The values of $\rho$ are chosen to be 0.0, 0.5, 1.0, and 2.0. We can observe that, compared with the performance of $\rho =0$, the robustness of $\rho \neq 0$ all increases. And $\rho=1$ achieves the best performance among the choices. Besides, we compare the performance of SNN with only LIF neurons. When $\rho$ increases, the clean accuracy goes down. However, DLIF SNN almost remains the same. Increasing $\rho$ also improves robustness with LIF, but not surpassing robustness with DLIF. Besides, we also test training with spike distances of TASAD and STD introduced in Section~\ref{sec:sensitivity}. Training with TASAD or STD is not as effective at increasing robustness as training with MS-MPPD.

\begin{figure}[h]
    \centering
\setcounter{subfigure}{0}
    \subfigure[Accuracy, AT]{ 
\includegraphics[width=0.45\linewidth]{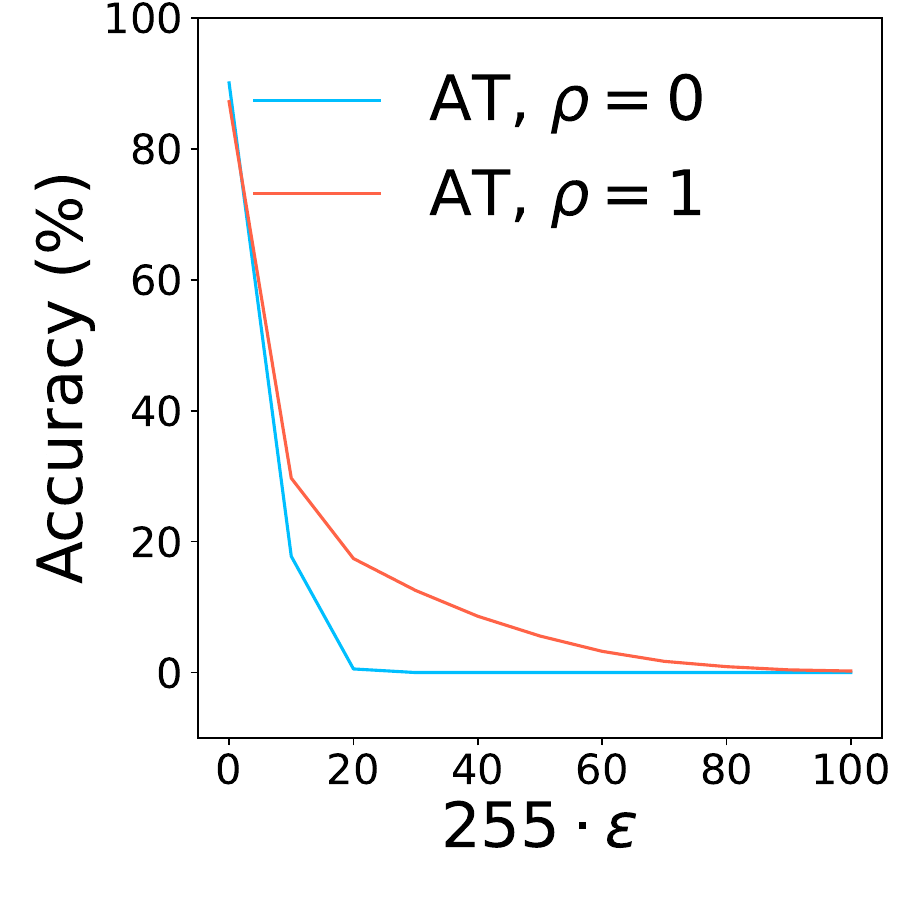}}
\subfigure[Accuracy, AT+Reg]{
\includegraphics[width=0.45\linewidth]{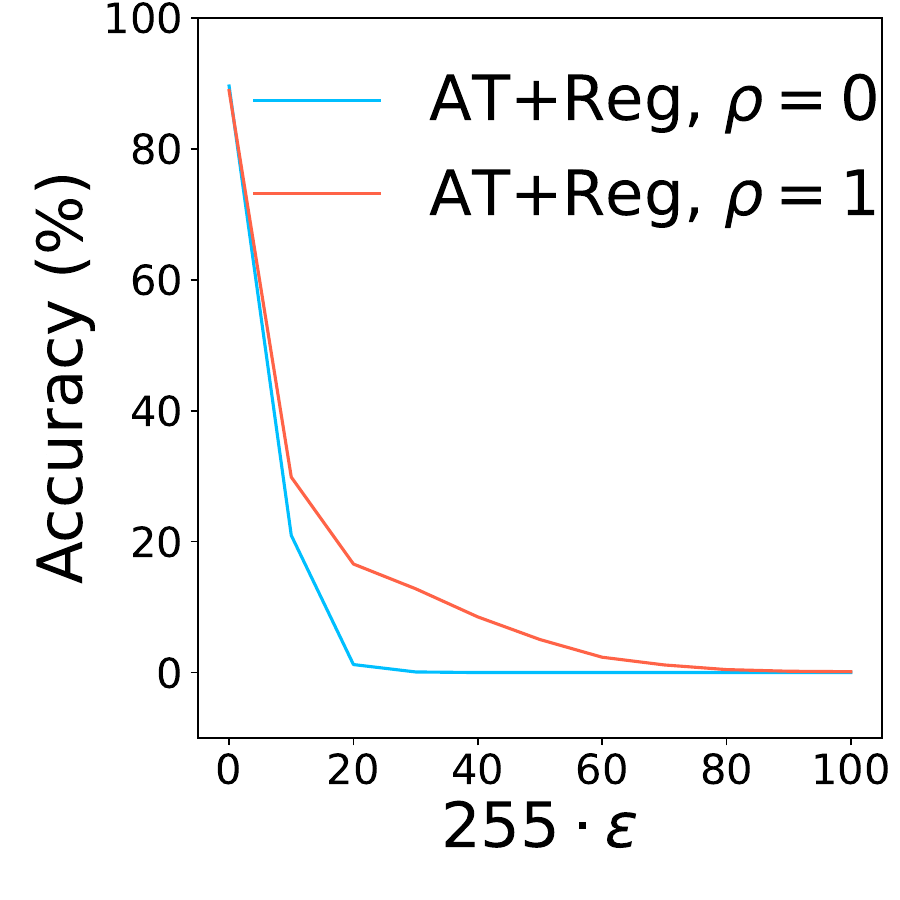}}

\subfigure[MS-MPPD, AT]{ 
\includegraphics[width=0.45\linewidth]{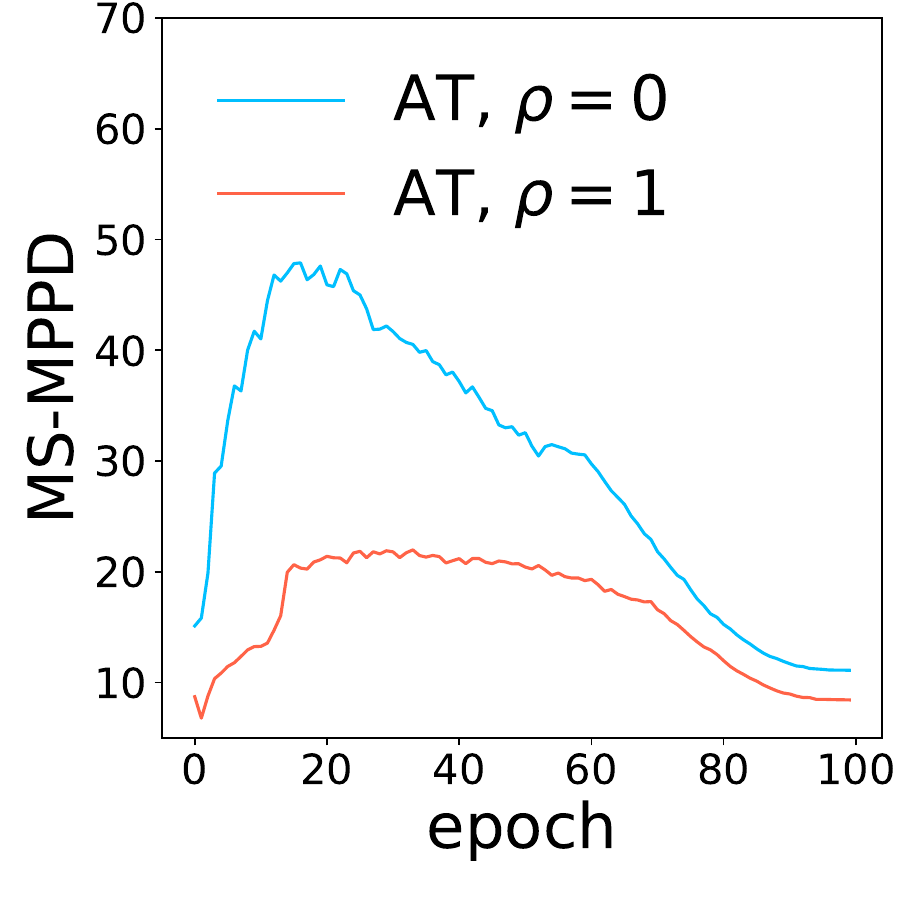}}
\subfigure[MS-MPPD, AT+Reg]{
\includegraphics[width=0.45\linewidth]{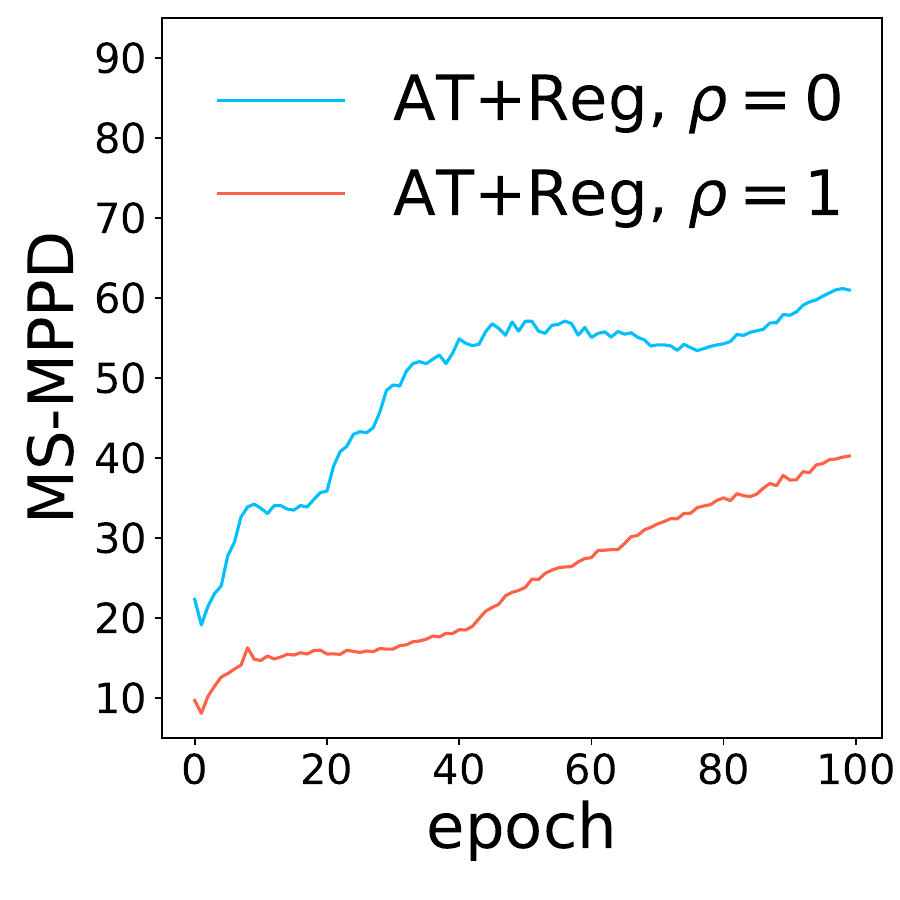}}
    \caption{Effect of the parameter $\rho$. }
    \label{fig:eps}
\end{figure}

We plot the trend of PGD$^{10}$ accuracy with attack intensity increasing on VGG11 trained with CIFAR-10 in Figure~\ref{fig:eps}(a)(b). The curve decreases slowly when $\rho=1$ compared with $\rho=0$, either with regularizer or not. The values of MS-MPPD are also constrained when $\rho=1$ (Figure~\ref{fig:eps}(c)(d)).

\section{Conclusions}

In this paper, we first give a perturbation metric from the viewpoint of neuronal dynamics. The perturbed input can lead to perturbation dynamics, which accurately represent the impact of perturbation. Our theoretical observations on the stability inspire us to propose a framework to improve the robustness of SNN with the assistance of a modified neuron and the mean square of the membrane potential perturbation dynamics. The experimental results show that our network exceeds the current SOTA methods for improving the robustness of SNN. Overall, we believe our work will increase the confidence of neuromorphic deployments in future safety-critical applications.

\section*{Acknowledgements}

This work was supported by the National Natural Science Foundation of China (62176003, 62088102) and by Beijing Nova Program (20230484362).

\section*{Impact Statement}

Our research is on how to build secure and robust spiking neural networks. There is no apparent negative social impact. Our proposed strategy enhances adversarial robustness, resulting in a considerably more favorable societal impact. Neuromorphic computing is receiving widespread attention, and it is crucial to build safe and stable spiking neural networks. We believe that our work can help the community focus on potential security threats of spiking neural networks in safety-critical applications.

\nocite{langley00}

\bibliography{icml}

\begin{thebibliography}{54}
\providecommand{\natexlab}[1]{#1}
\providecommand{\url}[1]{\texttt{#1}}
\expandafter\ifx\csname urlstyle\endcsname\relax
  \providecommand{\doi}[1]{doi: #1}\else
  \providecommand{\doi}{doi: \begingroup \urlstyle{rm}\Url}\fi

\bibitem[Bing et~al.(2018)Bing, Meschede, R{\"o}hrbein, Huang, and Knoll]{bing2018survey}
Bing, Z., Meschede, C., R{\"o}hrbein, F., Huang, K., and Knoll, A.~C.
\newblock A survey of robotics control based on learning-inspired spiking neural networks.
\newblock \emph{Frontiers in Neurorobotics}, 12:\penalty0 35, 2018.

\bibitem[Bu et~al.(2023)Bu, Ding, Hao, and Yu]{bu2023rate}
Bu, T., Ding, J., Hao, Z., and Yu, Z.
\newblock Rate gradient approximation attack threats deep spiking neural networks.
\newblock In \emph{Proceedings of the IEEE/CVF International Conference on Computer Vision}, pp.\  7896--7906, 2023.

\bibitem[Chen et~al.(2018)Chen, Rubanova, Bettencourt, and Duvenaud]{chen2018neural}
Chen, R.~T., Rubanova, Y., Bettencourt, J., and Duvenaud, D.~K.
\newblock Neural ordinary differential equations.
\newblock \emph{Advances in Neural Information Processing Systems}, 31, 2018.

\bibitem[Croce \& Hein(2020)Croce and Hein]{croce2020reliable}
Croce, F. and Hein, M.
\newblock Reliable evaluation of adversarial robustness with an ensemble of diverse parameter-free attacks.
\newblock In \emph{Proceedings of the 38th International Conference on Machine Learning}, pp.\  2206--2216. PMLR, 2020.

\bibitem[Davies et~al.(2018)Davies, Srinivasa, Lin, Chinya, Cao, Choday, Dimou, Joshi, Imam, Jain, et~al.]{davies2018loihi}
Davies, M., Srinivasa, N., Lin, T.-H., Chinya, G., Cao, Y., Choday, S.~H., Dimou, G., Joshi, P., Imam, N., Jain, S., et~al.
\newblock Loihi: {A} neuromorphic manycore processor with on-chip learning.
\newblock \emph{IEEE Micro}, 38\penalty0 (1):\penalty0 82--99, 2018.

\bibitem[DeBole et~al.(2019)DeBole, Taba, Amir, Akopyan, Andreopoulos, Risk, Kusnitz, Otero, Nayak, Appuswamy, et~al.]{debole2019truenorth}
DeBole, M.~V., Taba, B., Amir, A., Akopyan, F., Andreopoulos, A., Risk, W.~P., Kusnitz, J., Otero, C.~O., Nayak, T.~K., Appuswamy, R., et~al.
\newblock {T}rue{N}orth: Accelerating from zero to 64 million neurons in 10 years.
\newblock \emph{Computer}, 52\penalty0 (5):\penalty0 20--29, 2019.

\bibitem[Deng et~al.(2021)Deng, Li, Zhang, and Gu]{deng2021temporal}
Deng, S., Li, Y., Zhang, S., and Gu, S.
\newblock Temporal efficient training of spiking neural network via gradient re-weighting.
\newblock In \emph{International Conference on Learning Representations}, 2021.

\bibitem[Ding et~al.(2022)Ding, Bu, Yu, Huang, and Liu]{ding2022snn}
Ding, J., Bu, T., Yu, Z., Huang, T., and Liu, J.
\newblock {SNN-RAT}: Robustness-enhanced spiking neural network through regularized adversarial training.
\newblock \emph{Advances in Neural Information Processing Systems}, 35:\penalty0 24780--24793, 2022.

\bibitem[Ding et~al.(2024)Ding, Yu, Huang, and Liu]{ding2024enhancing}
Ding, J., Yu, Z., Huang, T., and Liu, J.~K.
\newblock Enhancing the robustness of spiking neural networks with stochastic gating mechanisms.
\newblock In \emph{Proceedings of the AAAI Conference on Artificial Intelligence}, volume~38, pp.\  492--502, 2024.

\bibitem[El-Allami et~al.(2021)El-Allami, Marchisio, Shafique, and Alouani]{el2021securing}
El-Allami, R., Marchisio, A., Shafique, M., and Alouani, I.
\newblock Securing deep spiking neural networks against adversarial attacks through inherent structural parameters.
\newblock In \emph{Design, Automation {\&} Test in Europe Conference {\&} Exhibition}, pp.\  774--779, 2021.

\bibitem[Fang et~al.(2020)Fang, Zhang, Yan, and Tang]{fang2020spike}
Fang, B., Zhang, Y., Yan, R., and Tang, H.
\newblock Spike trains encoding optimization for spiking neural networks implementation in fpga.
\newblock In \emph{International Conference on Advanced Computational Intelligence}, pp.\  412--418, 2020.

\bibitem[Fang et~al.(2021)Fang, Yu, Chen, Masquelier, Huang, and Tian]{fang2021incorporating}
Fang, W., Yu, Z., Chen, Y., Masquelier, T., Huang, T., and Tian, Y.
\newblock Incorporating learnable membrane time constant to enhance learning of spiking neural networks.
\newblock In \emph{Proceedings of the IEEE/CVF International Conference on Computer Vision}, pp.\  2661--2671, 2021.

\bibitem[Gerstner et~al.(2014)Gerstner, Kistler, Naud, and Paninski]{gerstner2014neuronal}
Gerstner, W., Kistler, W.~M., Naud, R., and Paninski, L.
\newblock \emph{{Neuronal} dynamics: {From} single neurons to networks and models of cognition}.
\newblock Cambridge University Press, 2014.

\bibitem[Goodfellow et~al.(2015)Goodfellow, Shlens, and Szegedy]{goodfellow2014explaining}
Goodfellow, I.~J., Shlens, J., and Szegedy, C.
\newblock Explaining and harnessing adversarial examples.
\newblock \emph{International Conference on Learning Representations}, 2015.

\bibitem[Guo et~al.(2023)Guo, Liu, Chen, Zhang, Peng, Zhang, Huang, and Ma]{guo2023rmp}
Guo, Y., Liu, X., Chen, Y., Zhang, L., Peng, W., Zhang, Y., Huang, X., and Ma, Z.
\newblock {RMP}-loss: Regularizing membrane potential distribution for spiking neural networks.
\newblock In \emph{Proceedings of the IEEE/CVF International Conference on Computer Vision}, pp.\  17391--17401, 2023.

\bibitem[Hao et~al.(2024)Hao, Bu, Shi, Huang, Yu, and Huang]{hao2024threaten}
Hao, Z., Bu, T., Shi, X., Huang, Z., Yu, Z., and Huang, T.
\newblock Threaten spiking neural networks through combining rate and temporal information.
\newblock In \emph{International Conference on Learning Representations}, 2024.

\bibitem[Khalil(2002)]{khalil2002nonlinear}
Khalil, H.~K.
\newblock \emph{Nonlinear systems}.
\newblock Prentice-Hall, 2002.

\bibitem[Kheradpisheh \& Masquelier(2020)Kheradpisheh and Masquelier]{kheradpisheh2020temporal}
Kheradpisheh, S.~R. and Masquelier, T.
\newblock Temporal backpropagation for spiking neural networks with one spike per neuron.
\newblock \emph{International Journal of Neural Systems}, 30\penalty0 (06):\penalty0 2050027, 2020.

\bibitem[Kim et~al.(2020)Kim, Park, Na, and Yoon]{kim2020spiking}
Kim, S., Park, S., Na, B., and Yoon, S.
\newblock Spiking-{YOLO}: Spiking neural network for energy-efficient object detection.
\newblock In \emph{Proceedings of the AAAI Conference on Artificial Intelligence}, pp.\  11270--11277, 2020.

\bibitem[Kim \& Panda(2021)Kim and Panda]{kim2021revisiting}
Kim, Y. and Panda, P.
\newblock Revisiting batch normalization for training low-latency deep spiking neural networks from scratch.
\newblock \emph{Frontiers in Neuroscience}, 15:\penalty0 773954--773954, 2021.

\bibitem[Kim et~al.(2022)Kim, Park, Moitra, Bhattacharjee, Venkatesha, and Panda]{kim2022rate}
Kim, Y., Park, H., Moitra, A., Bhattacharjee, A., Venkatesha, Y., and Panda, P.
\newblock Rate coding or direct coding: Which one is better for accurate, robust, and energy-efficient spiking neural networks?
\newblock In \emph{IEEE International Conference on Acoustics, Speech, and Signal Processing}, pp.\  71--75, 2022.

\bibitem[Kim et~al.(2023)Kim, Li, Park, Venkatesha, Hambitzer, and Panda]{kim2023exploring}
Kim, Y., Li, Y., Park, H., Venkatesha, Y., Hambitzer, A., and Panda, P.
\newblock Exploring temporal information dynamics in spiking neural networks.
\newblock In \emph{Proceedings of the AAAI Conference on Artificial Intelligence}, volume~37, pp.\  8308--8316, 2023.

\bibitem[Kojima \& Okamoto(2022)Kojima and Okamoto]{kojima2022learning}
Kojima, R. and Okamoto, Y.
\newblock Learning deep input-output stable dynamics.
\newblock \emph{Advances in Neural Information Processing Systems}, 35:\penalty0 8187--8198, 2022.

\bibitem[Kundu et~al.(2021)Kundu, Pedram, and Beerel]{kundu2021hire}
Kundu, S., Pedram, M., and Beerel, P.~A.
\newblock {HIRE-SNN}: Harnessing the inherent robustness of energy-efficient deep spiking neural networks by training with crafted input noise.
\newblock In \emph{Proceedings of the IEEE/CVF International Conference on Computer Vision}, pp.\  5209--5218, 2021.

\bibitem[Kurakin et~al.(2017)Kurakin, Goodfellow, and Bengio]{kurakin2016adversarial}
Kurakin, A., Goodfellow, I.~J., and Bengio, S.
\newblock Adversarial examples in the physical world.
\newblock In \emph{International Conference on Learning Representations, Workshop Track Proceedings}, 2017.

\bibitem[Lawrence et~al.(2020)Lawrence, Loewen, Forbes, Backstrom, and Gopaluni]{lawrence2020almost}
Lawrence, N., Loewen, P., Forbes, M., Backstrom, J., and Gopaluni, B.
\newblock Almost surely stable deep dynamics.
\newblock \emph{Advances in Neural Information Processing Systems}, 33:\penalty0 18942--18953, 2020.

\bibitem[Liang et~al.(2022)Liang, Xu, Hu, Deng, and Xie]{liang2022toward}
Liang, L., Xu, K., Hu, X., Deng, L., and Xie, Y.
\newblock Toward robust spiking neural network against adversarial perturbation.
\newblock In \emph{Advances in Neural Information Processing Systems}, volume~35, pp.\  10244--10256, 2022.

\bibitem[Liang et~al.(2023)Liang, Hu, Deng, Wu, Li, Ding, Li, and Xie]{liang2021exploring}
Liang, L., Hu, X., Deng, L., Wu, Y., Li, G., Ding, Y., Li, P., and Xie, Y.
\newblock Exploring adversarial attack in spiking neural networks with spike-compatible gradient.
\newblock \emph{IEEE Transactions on Neural Networks and Learning Systems}, 34\penalty0 (5):\penalty0 2569--2583, 2023.

\bibitem[Maass(1997)]{maas1997networks}
Maass, W.
\newblock {Networks} of spiking neurons: {The} third generation of neural network models.
\newblock \emph{Neural Networks}, 10\penalty0 (9):\penalty0 1659--1671, 1997.

\bibitem[Madry et~al.(2018)Madry, Makelov, Schmidt, Tsipras, and Vladu]{aleksander2018towards}
Madry, A., Makelov, A., Schmidt, L., Tsipras, D., and Vladu, A.
\newblock Towards deep learning models resistant to adversarial attacks.
\newblock In \emph{International Conference on Learning Representations}, 2018.

\bibitem[Marchisio et~al.(2021)Marchisio, Pira, Martina, Masera, and Shafique]{alberto2021dvs}
Marchisio, A., Pira, G., Martina, M., Masera, G., and Shafique, M.
\newblock {DVS-Attacks}: {Adversarial} attacks on dynamic vision sensors for spiking neural networks.
\newblock In \emph{International Joint Conference on Neural Networks}, pp.\  1--9, 2021.

\bibitem[Nieves \& Goodman(2021)Nieves and Goodman]{nicolas2021sparse}
Nieves, N.~P. and Goodman, D. F.~M.
\newblock Sparse spiking gradient descent.
\newblock In \emph{Advances in Neural Information Processing Systems}, pp.\  11795--11808, 2021.

\bibitem[{\"O}zdenizci \& Legenstein(2021){\"O}zdenizci and Legenstein]{ozdenizci2021training}
{\"O}zdenizci, O. and Legenstein, R.
\newblock Training adversarially robust sparse networks via bayesian connectivity sampling.
\newblock In \emph{Proceedings of the 38th International Conference on Machine Learning}, pp.\  8314--8324, 2021.

\bibitem[Pei et~al.(2019)Pei, Deng, Song, Zhao, Zhang, Wu, Wang, Zou, Wu, He, et~al.]{pei2019towards}
Pei, J., Deng, L., Song, S., Zhao, M., Zhang, Y., Wu, S., Wang, G., Zou, Z., Wu, Z., He, W., et~al.
\newblock Towards artificial general intelligence with hybrid {T}ianjic chip architecture.
\newblock \emph{Nature}, 572\penalty0 (7767):\penalty0 106--111, 2019.

\bibitem[Rathi \& Roy(2021)Rathi and Roy]{rathi2021diet}
Rathi, N. and Roy, K.
\newblock {DIET-SNN}: A low-latency spiking neural network with direct input encoding and leakage and threshold optimization.
\newblock \emph{IEEE Transactions on Neural Networks and Learning Systems}, 34\penalty0 (6):\penalty0 3174--3182, 2021.

\bibitem[Sharmin et~al.(2019)Sharmin, Panda, Sarwar, Lee, Ponghiran, and Roy]{sharmin2019comprehensive}
Sharmin, S., Panda, P., Sarwar, S.~S., Lee, C., Ponghiran, W., and Roy, K.
\newblock A comprehensive analysis on adversarial robustness of spiking neural networks.
\newblock In \emph{International Joint Conference on Neural Networks}, pp.\  1--8, 2019.

\bibitem[Sharmin et~al.(2020)Sharmin, Rathi, Panda, and Roy]{sharmin2020inherent}
Sharmin, S., Rathi, N., Panda, P., and Roy, K.
\newblock Inherent adversarial robustness of deep spiking neural networks: {Effects} of discrete input encoding and non-linear activations.
\newblock In \emph{European Conference on Computer Vision}, pp.\  399--414, 2020.

\bibitem[Shi et~al.(2024{\natexlab{a}})Shi, Ding, Hao, and Yu]{shi2024towards}
Shi, X., Ding, J., Hao, Z., and Yu, Z.
\newblock Towards energy efficient spiking neural networks: An unstructured pruning framework.
\newblock In \emph{International Conference on Learning Representations}, 2024{\natexlab{a}}.

\bibitem[Shi et~al.(2024{\natexlab{b}})Shi, Hao, and Yu]{shi2024spikingresformer}
Shi, X., Hao, Z., and Yu, Z.
\newblock Spikingresformer: Bridging resnet and vision transformer in spiking neural networks.
\newblock In \emph{Proceedings of the IEEE/CVF International Conference on Computer Vision}, 2024{\natexlab{b}}.

\bibitem[Szegedy et~al.(2014)Szegedy, Zaremba, Sutskever, Bruna, Erhan, Goodfellow, and Fergus]{szegedy2013intriguing}
Szegedy, C., Zaremba, W., Sutskever, I., Bruna, J., Erhan, D., Goodfellow, I.~J., and Fergus, R.
\newblock Intriguing properties of neural networks.
\newblock In \emph{International Conference on Learning Representations}, 2014.

\bibitem[Wang et~al.(2019)Wang, Zou, Yi, Bailey, Ma, and Gu]{wang2019improving}
Wang, Y., Zou, D., Yi, J., Bailey, J., Ma, X., and Gu, Q.
\newblock Improving adversarial robustness requires revisiting misclassified examples.
\newblock In \emph{International Conference on Learning Representations}, 2019.

\bibitem[Wong et~al.(2019)Wong, Rice, and Kolter]{wong2019fast}
Wong, E., Rice, L., and Kolter, J.~Z.
\newblock Fast is better than free: Revisiting adversarial training.
\newblock In \emph{International Conference on Learning Representations}, 2019.

\bibitem[Wu et~al.(2018)Wu, Deng, Li, Zhu, and Shi]{wu2018spatio}
Wu, Y., Deng, L., Li, G., Zhu, J., and Shi, L.
\newblock Spatio-temporal backpropagation for training high-performance spiking neural networks.
\newblock \emph{Frontiers in Neuroscience}, 12:\penalty0 331, 2018.

\bibitem[Xu et~al.(2022)Xu, Li, Shen, Zhang, Liu, Tang, and Pan]{xu2022hierarchical}
Xu, Q., Li, Y., Shen, J., Zhang, P., Liu, J.~K., Tang, H., and Pan, G.
\newblock Hierarchical spiking-based model for efficient image classification with enhanced feature extraction and encoding.
\newblock \emph{IEEE Transactions on Neural Networks and Learning Systems}, 2022.

\bibitem[Xu et~al.(2023)Xu, Li, Shen, Liu, Tang, and Pan]{xu2023constructing}
Xu, Q., Li, Y., Shen, J., Liu, J.~K., Tang, H., and Pan, G.
\newblock Constructing deep spiking neural networks from artificial neural networks with knowledge distillation.
\newblock In \emph{Proceedings of the IEEE/CVF International Conference on Computer Vision}, pp.\  7886--7895, 2023.

\bibitem[Xu et~al.(2024)Xu, Gao, Shen, Li, Ran, Tang, and Pan]{xu2024enhancing}
Xu, Q., Gao, Y., Shen, J., Li, Y., Ran, X., Tang, H., and Pan, G.
\newblock Enhancing adaptive history reserving by spiking convolutional block attention module in recurrent neural networks.
\newblock \emph{Advances in Neural Information Processing Systems}, 36, 2024.

\bibitem[Yamazaki et~al.(2022)Yamazaki, Vo-Ho, Bulsara, and Le]{yamazaki2022spiking}
Yamazaki, K., Vo-Ho, V.-K., Bulsara, D., and Le, N.
\newblock Spiking neural networks and their applications: A review.
\newblock \emph{Brain Sciences}, 12\penalty0 (7):\penalty0 863, 2022.

\bibitem[Yao et~al.(2022)Yao, Li, Mo, and Cheng]{yao2022glif}
Yao, X., Li, F., Mo, Z., and Cheng, J.
\newblock {GLIF}: A unified gated leaky integrate-and-fire neuron for spiking neural networks.
\newblock \emph{Advances in Neural Information Processing Systems}, 35:\penalty0 32160--32171, 2022.

\bibitem[Zenke et~al.(2021)Zenke, Boht{\'e}, Clopath, Com{\c{s}}a, G{\"o}ltz, Maass, Masquelier, Naud, Neftci, Petrovici, et~al.]{zenke2021visualizing}
Zenke, F., Boht{\'e}, S.~M., Clopath, C., Com{\c{s}}a, I.~M., G{\"o}ltz, J., Maass, W., Masquelier, T., Naud, R., Neftci, E.~O., Petrovici, M.~A., et~al.
\newblock Visualizing a joint future of neuroscience and neuromorphic engineering.
\newblock \emph{Neuron}, 109\penalty0 (4):\penalty0 571--575, 2021.

\bibitem[Zhang et~al.(2018)Zhang, Cisse, Dauphin, and Lopez-Paz]{zhang2018mixup}
Zhang, H., Cisse, M., Dauphin, Y.~N., and Lopez-Paz, D.
\newblock mixup: Beyond empirical risk minimization.
\newblock In \emph{International Conference on Learning Representations}, 2018.

\bibitem[Zhang et~al.(2019)Zhang, Yu, Jiao, Xing, El~Ghaoui, and Jordan]{zhang2019theoretically}
Zhang, H., Yu, Y., Jiao, J., Xing, E., El~Ghaoui, L., and Jordan, M.
\newblock Theoretically principled trade-off between robustness and accuracy.
\newblock In \emph{Proceedings of the 38th International Conference on Machine Learning}, pp.\  7472--7482. PMLR, 2019.

\bibitem[Zhang et~al.(2022)Zhang, Wang, Wu, Belatreche, Amornpaisannon, Zhang, Miriyala, Qu, Chua, Carlson, et~al.]{zhang2022rectified}
Zhang, M., Wang, J., Wu, J., Belatreche, A., Amornpaisannon, B., Zhang, Z., Miriyala, V. P.~K., Qu, H., Chua, Y., Carlson, T.~E., et~al.
\newblock Rectified linear postsynaptic potential function for backpropagation in deep spiking neural networks.
\newblock \emph{IEEE Transactions on Neural Networks and Learning Systems}, 33\penalty0 (5):\penalty0 1947--1958, 2022.

\bibitem[Zhang \& Li(2020)Zhang and Li]{zhang2020temporal}
Zhang, W. and Li, P.
\newblock Temporal spike sequence learning via backpropagation for deep spiking neural networks.
\newblock \emph{Advances in Neural Information Processing Systems}, pp.\  12022--12033, 2020.

\bibitem[Zhu et~al.(2024)Zhu, Ding, Huang, Xie, and Yu]{zhu2024online}
Zhu, Y., Ding, J., Huang, T., Xie, X., and Yu, Z.
\newblock Online stabilization of spiking neural networks.
\newblock In \emph{International Conference on Learning Representations}, 2024.

\end{thebibliography}
\bibliographystyle{icml2024}

\newpage
\appendix
\onecolumn
\section{Proofs}

In this section, we are going to proof Theorem~\ref{thm:stability} in the main text.


\begin{proof}
SNN with LIF neuron infer multiple discrete time steps ($T$ steps) to get the output. We first here transform the $L_2$ stability of a nonlinear continuous-time system to a discrete-time system. Given signal $\boldsymbol{x}$ with discrete temporal axis, the $L_2$ norm is defined as $\left\| \boldsymbol{x}_{[:T]} \right\| _{L_2}=\sqrt{\sum\nolimits_{t=0}^T{\left\| \boldsymbol{x}\left[ t \right] \right\| ^2}}
$.

Then for the membrane potential perturbation dynamics $\boldsymbol{\varepsilon }^l[ t ] =\lambda \boldsymbol{\varepsilon }^l[ t-1 ] +\boldsymbol{W}^l\varDelta \boldsymbol{s}^{l-1}[ t ] $, our aim is to determine $\gamma^l$ and $\beta^l$ for the following formula:
\begin{equation}
    \left\| {\boldsymbol{\varepsilon }^l}_{[:T ]} \right\| _{L_2}\leq \gamma ^l\left\| \varDelta {\boldsymbol{s}^{l-1}}_{[:T ]} \right\| _{L_2}+\beta ^l,
    \label{eq:proof_1}
\end{equation}
where
\begin{align}
\left\| \boldsymbol{\varepsilon }^l_{[:T]} \right\| _{L_2}&=\sqrt{\sum\nolimits_{t=1}^T{\left\| \boldsymbol{\varepsilon }^l\left[ t \right] \right\| ^2}}, \label{eq:varepsilon} \\
\left\| \varDelta \boldsymbol{s}^{l-1}_{[:T]} \right\| _{L_2}&=\sqrt{\sum\nolimits_{t=1}^T{\left\| \varDelta \boldsymbol{s}^{l-1}\left[ t \right] \right\| ^2}}.
\end{align}
By iterating the perturbation dynamics, we can obtain
\begin{equation}
\boldsymbol{\varepsilon }^l\left[ t \right] =\boldsymbol{W}^l\varDelta \boldsymbol{s}^{l-1}\left[ t \right] +\lambda \boldsymbol{W}^l\varDelta \boldsymbol{s}^{l-1}\left[ t-1 \right] +\cdots +\lambda ^{t-1}\boldsymbol{W}^l\varDelta \boldsymbol{s}^{l-1}\left[ 1 \right] ,
\end{equation}
Thus, according to the inequality of norm, we have
\begin{align}
\left\| \boldsymbol{\varepsilon }^l\left[ t \right] \right\| _2&\leq \left\| \boldsymbol{W}^l\varDelta \boldsymbol{s}^{l-1}\left[ t \right] \right\| _2+\lambda \left\| \boldsymbol{W}^l\varDelta \boldsymbol{s}^{l-1}\left[ t-1 \right] \right\| _2+\cdots +\lambda ^{t-1}\left\| \boldsymbol{W}^l\varDelta \boldsymbol{s}^{l-1}\left[ 1 \right] \right\| _2 \\
&\leq \left\| \boldsymbol{W}^l \right\| \left( \left\| \varDelta \boldsymbol{s}^{l-1}\left[ t \right] \right\| _2+\lambda \left\| \varDelta \boldsymbol{s}^{l-1}\left[ t-1 \right] \right\| _2+\cdots +\lambda ^{t-1}\left\| \varDelta \boldsymbol{s}^{l-1}\left[ 1 \right] \right\| _2 \right) ,
\end{align}
where $ \left\| \boldsymbol{W}^l \right\|$ is the spectral norm of the weight.
Therefore, we can reformulate Eq.~\ref{eq:varepsilon} into the following:
\begin{align}
\left\| \boldsymbol{\varepsilon }^l_{[:T]} \right\| _{L_2}&=\sqrt{\sum\nolimits_{t=1}^T{\left\| \boldsymbol{\varepsilon }^l\left[ t \right] \right\| ^2}}
\\
&\leqslant \left\| \boldsymbol{W}^l \right\| \sqrt{1\cdot \left\| \varDelta \boldsymbol{s}^{l-1}\left[ T \right] \right\| ^2+\cdots +\left( 1+\lambda +\cdots +\lambda ^{T-1} \right) \left\| \varDelta \boldsymbol{s}^{l-1}\left[ 1 \right] \right\| ^2}
\\
&\leqslant \left\| \boldsymbol{W}^l \right\| \sqrt{1\cdot \left\| \varDelta \boldsymbol{s}^{l-1}\left[ T \right] \right\| ^2+\cdots +\left( 1-\lambda ^T \right) /\left( 1-\lambda \right) \left\| \varDelta \boldsymbol{s}^{l-1}\left[ 1 \right] \right\| ^2}
\\
&\leqslant \sqrt{1/(1-\lambda)}\left\| \boldsymbol{W}^l \right\| \sqrt{\sum\nolimits_{t=1}^T{\left\| \varDelta \boldsymbol{s}^{l-1}\left[ t \right] \right\| ^2}}
\\
&\leqslant \sqrt{1/(1-\lambda)}\left\| \boldsymbol{W}^l \right\| \left\| \varDelta \boldsymbol{s}^{l-1} \right\| _{L_2}.
\end{align}
Hence, $\gamma^l=\sqrt{1/(1-\lambda)} \Vert \boldsymbol{W}^l \Vert$ and $\beta^l=0$. $\Vert \boldsymbol{W}^l \Vert$ is the spectral norm of the weight.

\end{proof}

\section{Effect of DLIF neuron}

\begin{figure*}[h]
    \centering
\setcounter{subfigure}{0}
    \subfigure[$\rho=0.0$]{
\includegraphics[width=0.45\linewidth]{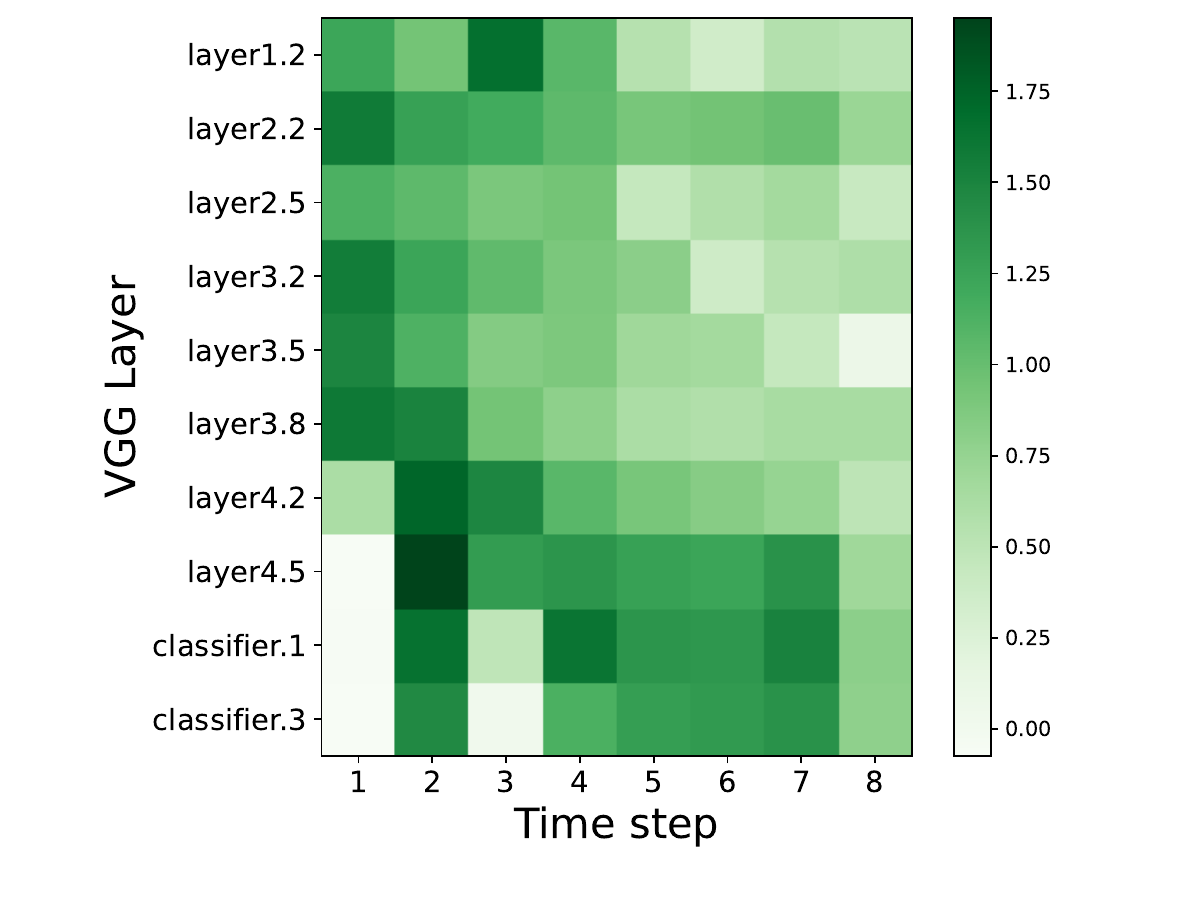}}
\subfigure[$\rho=1.0$]{
\includegraphics[width=0.45\linewidth]{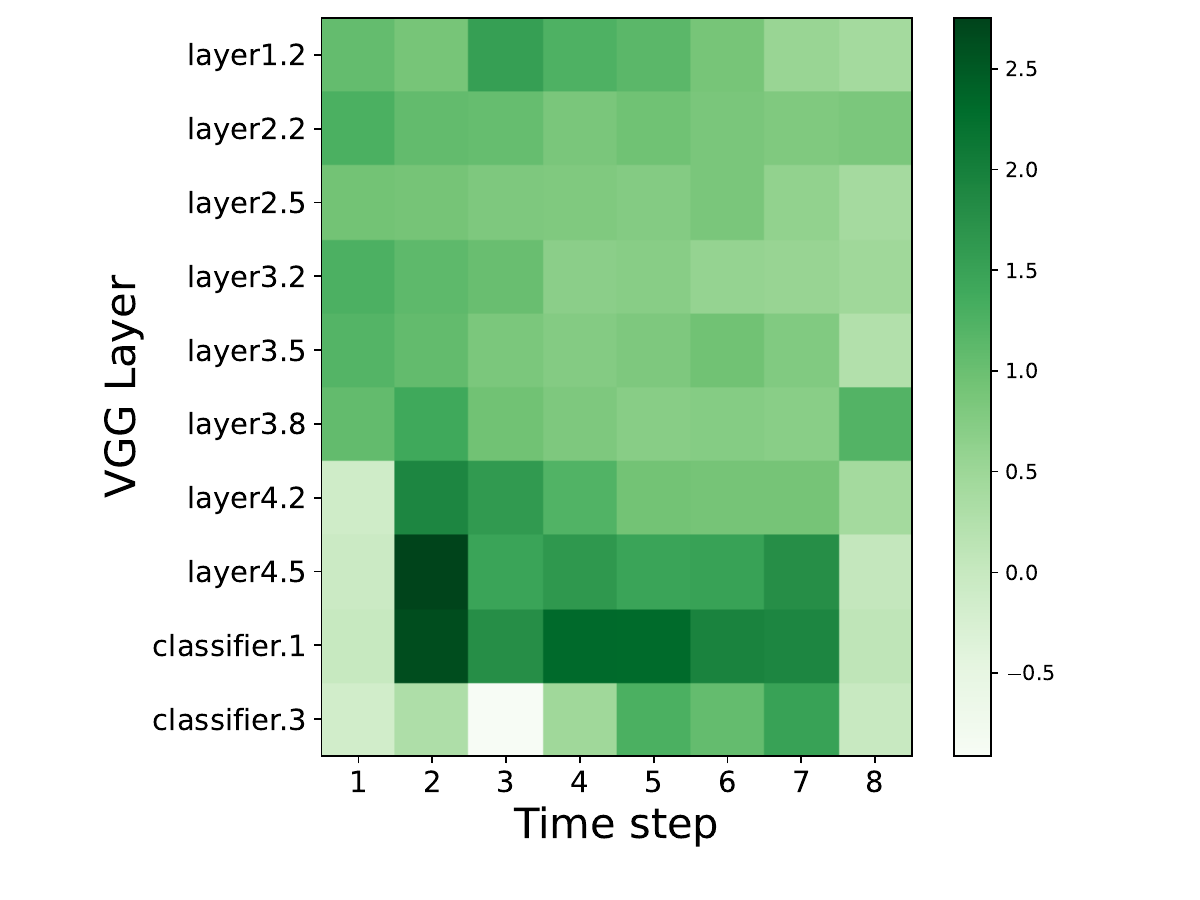}}

\subfigure[Average across time steps]{
\includegraphics[width=0.45\linewidth]{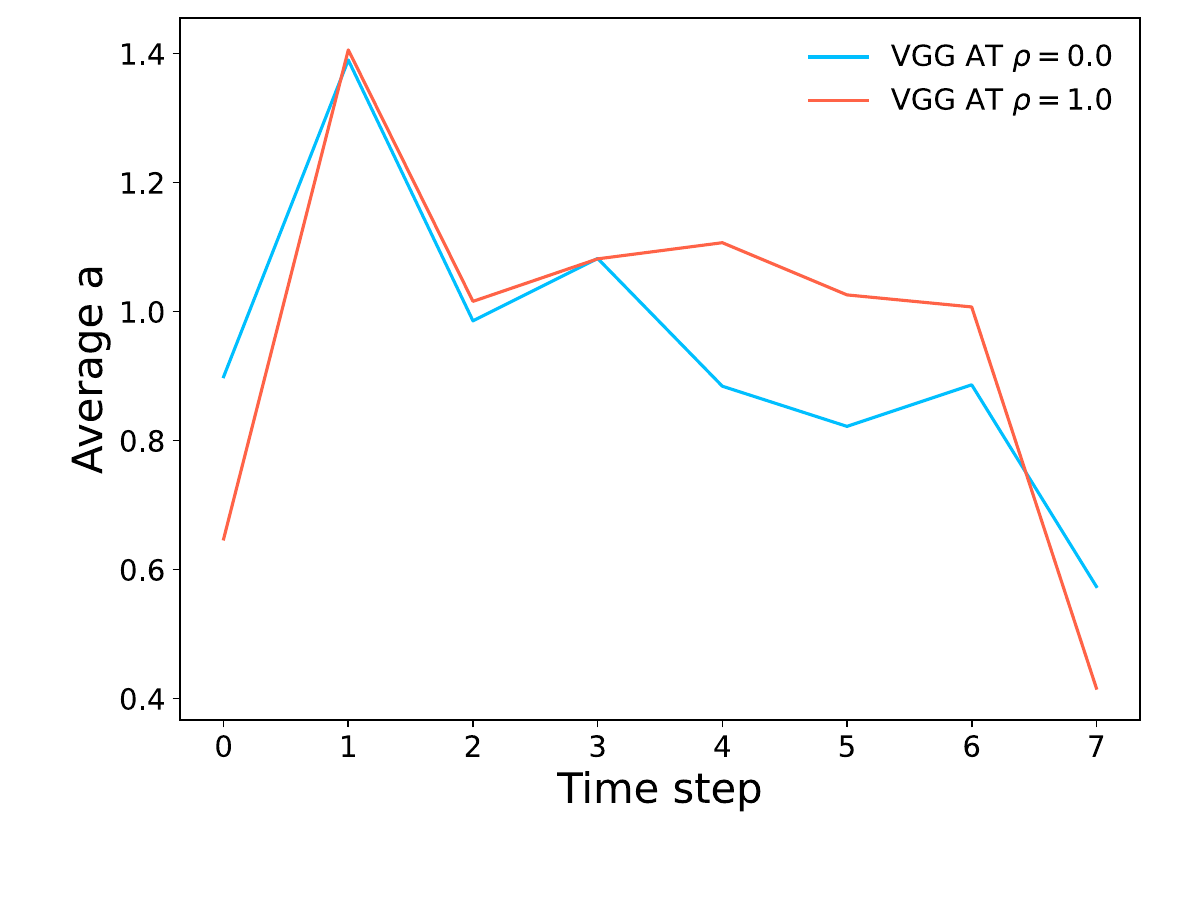}}
\subfigure[Average across layers]{
\includegraphics[width=0.45\linewidth]{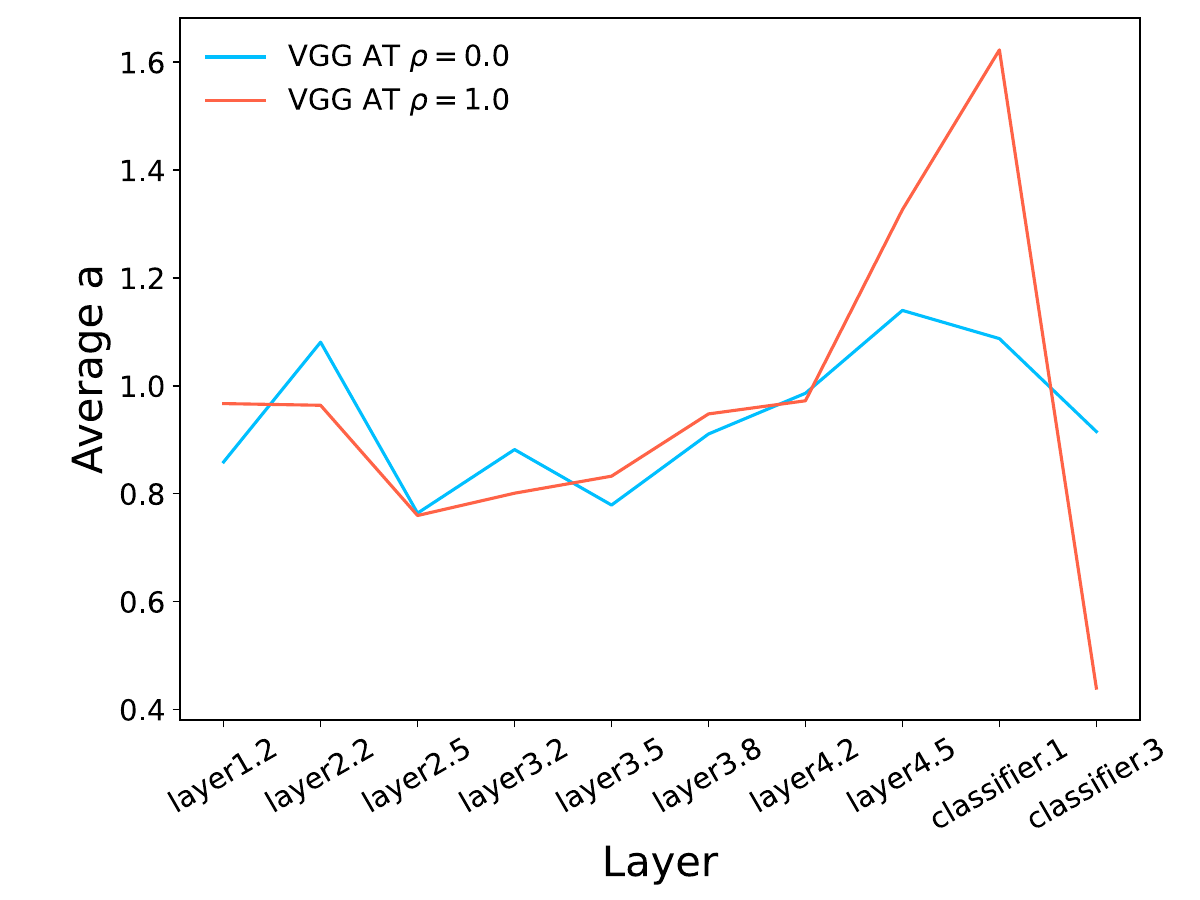}}
    \caption{Visualization of parameters in DLIF after proper training.}
    \label{fig:dlif}
\end{figure*}

We propose to use DLIF with varying trainable parameters to take the place of the vanilla LIF neuron. We would like to visualize the effect of DLIF. Note that the vanilla LIF neuron can be seen as a DLIF that fixes its trainable parameter to be 1.0. We use our VGG11 AT model on CIFAR-10 trained with $\rho=1$ and $\rho=0$ to visualize the parameters. The results are shown in Figure~\ref{fig:dlif}(a)(b). We also calculate their average values across time steps or across layers in Figure~\ref{fig:dlif}(c)(d). We can see that when $\rho=1$, the overall parameters are larger than the parameters when $\rho=1$. This trend is more obvious when the number of layers deepens or when the time step increases. Generally speaking, the parameters after training do not deviate too far from their initial values, and the values of these parameters are all near 1.

\begin{figure*}[t]
    \centering 
\setcounter{subfigure}{0}
    \subfigure[Task loss, AT]{
\includegraphics[width=0.24\linewidth]{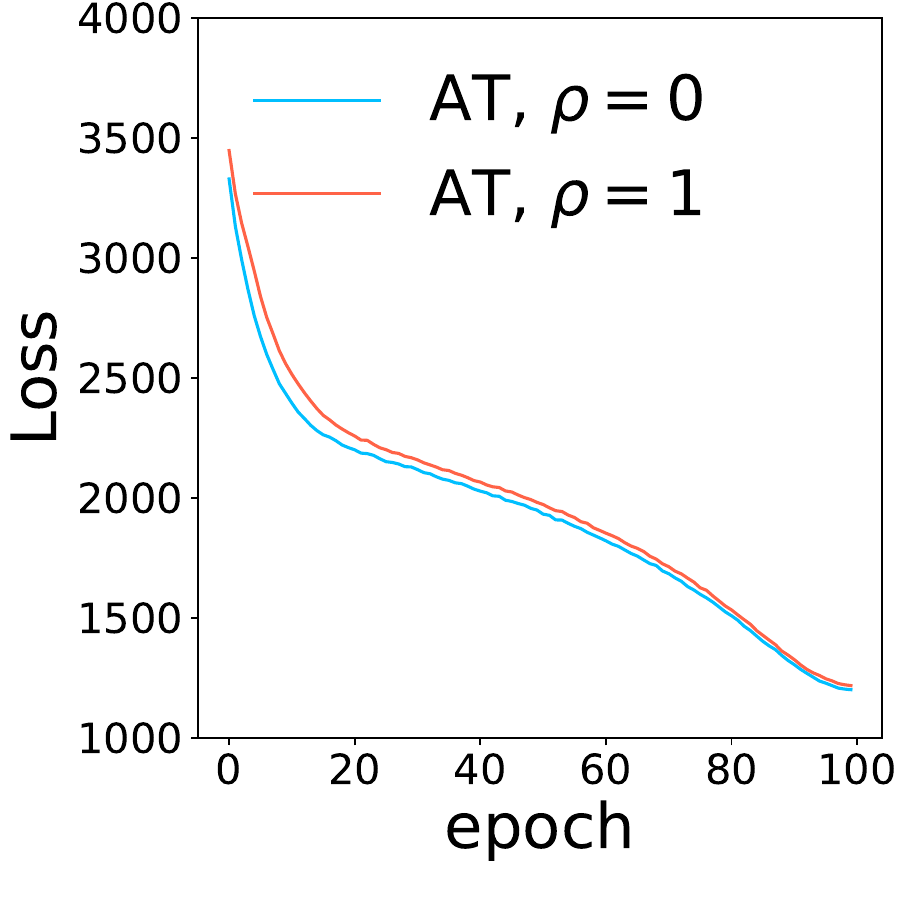}}
\subfigure[Task loss, AT+Reg]{
\includegraphics[width=0.24\linewidth]{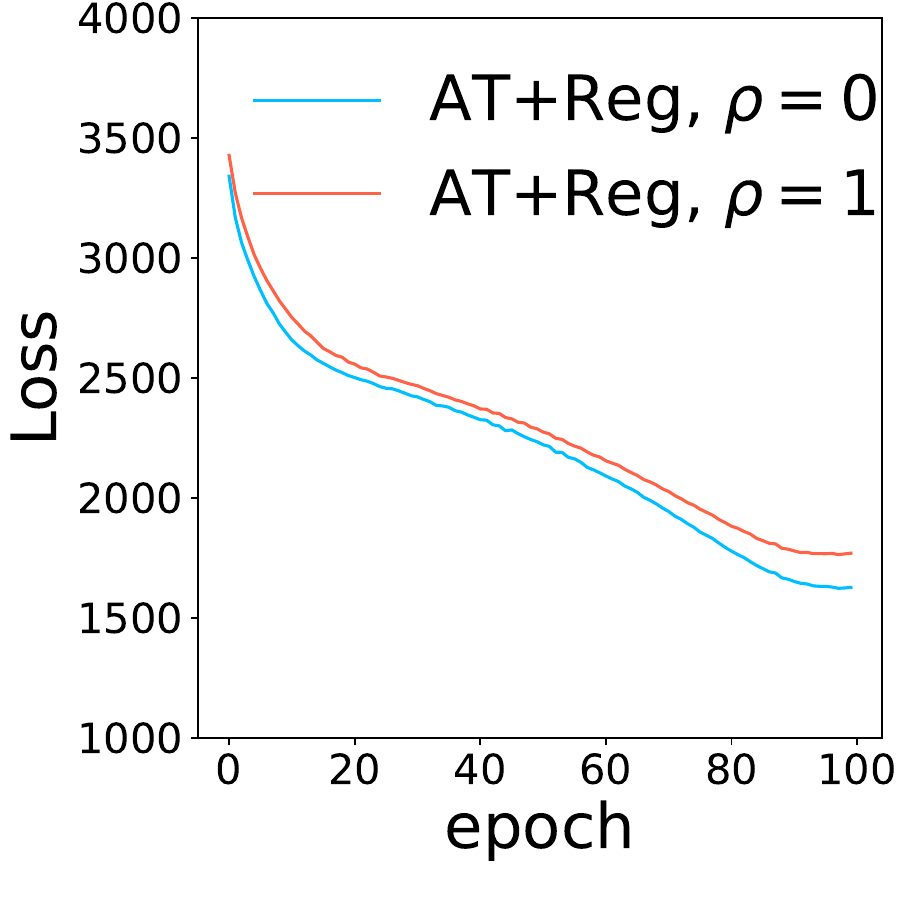}}
\subfigure[MS-MPPD, AT]{
\includegraphics[width=0.24\linewidth]{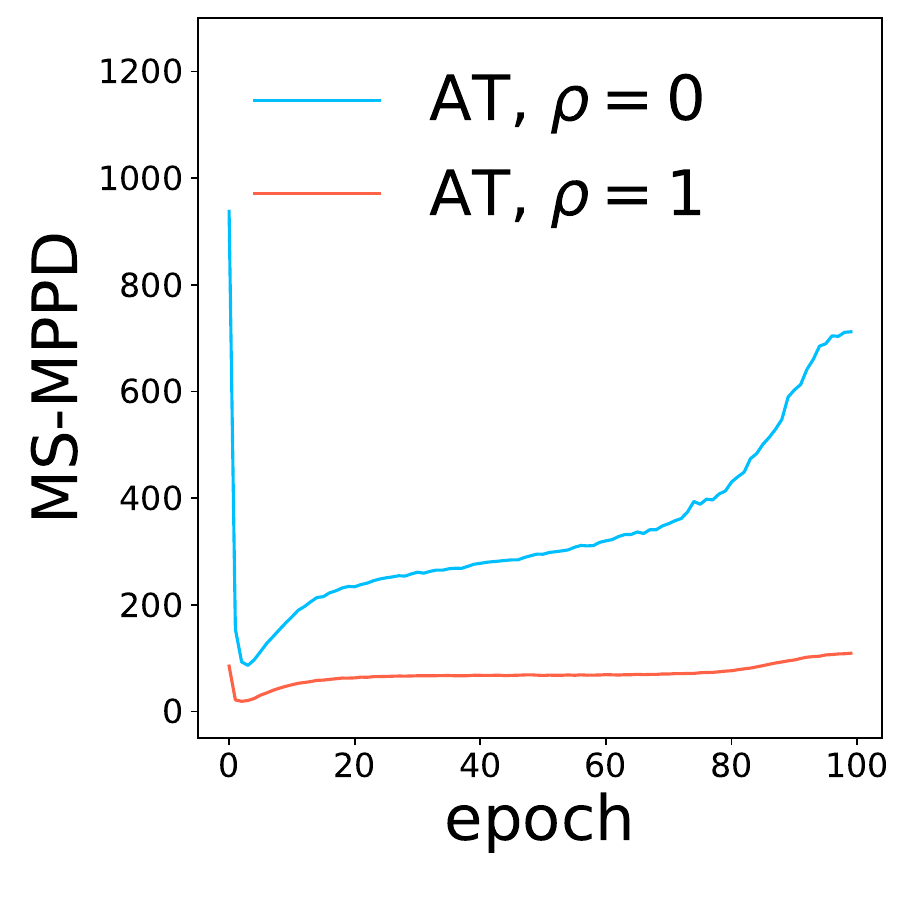}}
\subfigure[MS-MPPD, AT+Reg]{
\includegraphics[width=0.24\linewidth]{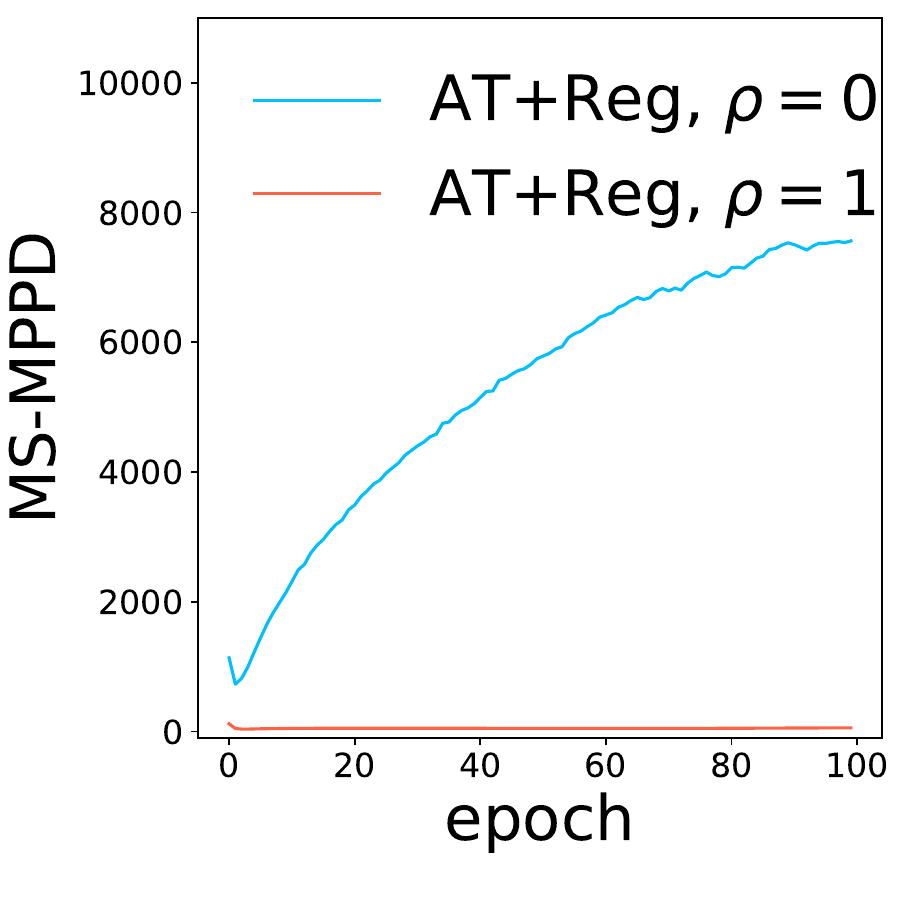}}
    \caption{Visualization of training process of WRN16.}
    \label{fig:wrn_loss}
\end{figure*}

\section{Implementation Details}

We conduct experiments on classification tasks on the CIFAR-10 and CIFAR-100 datasets. The SNN architectures used are VGG11 and WRN16. SNN uses direct encoding, and the encoding step size is 8. The number of training epochs is 100. The batch size is 64. We used float16 floating point precision during training. We use the SGD optimizer with an initial learning rate of 0.1. During training, the learning rate will decay to 0 in a cosine manner. The leakage factor for all SNNs is equal to 0.99. For models without regularization, we add l2 regularization terms with an intensity of 0.0005 during the model training process. 

We utilize SNN versions of the VGG5 and VGG11 networks, tailored for 32 $\times$ 32 image input. We've chosen these architectures for comparative analysis against three related works: \cite{sharmin2020inherent} (VGG5 and ResNet20 for CIFAR10, VGG11 for CIFAR100); \cite{kundu2021hire} (VGG5 and ResNet12 for CIFAR10, VGG11 and ResNet12 for CIFAR100); and \cite{ding2022snn} (current SOTA, VGG11, and WRN16 for both CIFAR10 and CIFAR100). Thus, we selected VGG11 and WRN16 for CIFAR10 and CIFAR100. For models using the regularizer in SNN-RAT, we also set the penalty intensity separately. For CIFAR-10, we set the intensity of the VGG11 model to 0.0005 and the intensity of the WRN16 model to 0.004; for CIFAR-100, we set the intensity of the VGG11 model to 0.001 and the intensity of the WRN16 model to 0.004.

Based on the above settings, we visualized the training process of CIFAR-100 WRN16. Our results can be seen in Figure~\ref{fig:wrn_loss}. During the training process, we saved the changes in $\mathcal{L}_{task}$ and MS-MPPD of this model. We find that when $\rho=1$ is used, $\mathcal{L}_{task}$ will increase compared to $\rho=0$, and the corresponding MS-MPPD will decrease. This shows that reducing MS-MPPD during training is similar to adding a regularizer. The above phenomenon is even more pronounced when used with other regularizations.

We performed adversarial attacks on spiking neural networks following previous literature \cite{kundu2021hire,ding2022snn}. First, we identify misclassification as the attacker's goal. By unfolding the dynamics of LIF neurons and applying surrogate functions to the non-differentiable Heaviside function, the network is able to backpropagate the gradient. Next, we perturbed the input in a direction that maximized the loss function using the computed gradient. We can employ FGSM and PGD as gradient-based attack methods.

We conduct experiments with our proposed training scheme with the regularizer proposed in \cite{ding2022snn} (RAT), as the two methods are orthogonal to each other. In RAT, the authors propose the use of spectral norm constraints on the weights, aiming to reduce the spike distance before and after the perturbation. In our work, we proposed to reduce the mean square of the membrane potential perturbation, and the implementation is to add a loss to the classification loss function. This does not conflict with the constraints on the weights, and the optimization goals of the two are consistent, which is to improve the robustness of SNN.

\end{document}